\documentclass[lettersize,journal]{IEEEtran}
\usepackage{amsmath,amsfonts}
\usepackage{algorithmic}
\usepackage{array}
\usepackage[caption=false,font=normalsize,labelfont=sf,textfont=sf]{subfig}
\usepackage{textcomp}
\usepackage{mathtools}
\usepackage{epsfig}
\usepackage{stfloats}
\usepackage{url}
\usepackage{verbatim}
\usepackage{times}
\usepackage{graphicx}
\usepackage{amssymb}

\usepackage[table]{xcolor}
\usepackage{gensymb}
\usepackage{multirow}
\usepackage{color}   
\usepackage{booktabs}
\definecolor{gray}{RGB}{229, 230, 230}
\def\360{360$\degree$}
\def\BibTeX{{\rm B\kern-.05em{\sc i\kern-.025em b}\kern-.08em
    T\kern-.1667em\lower.7ex\hbox{E}\kern-.125emX}}
\usepackage{balance}
\begin{document}
\title{SDGE: Stereo Guided Depth Estimation for 360\degree Camera Sets}
\author{
Jialei Xu,
Wei Yin,
Dong Gong,
Junjun~Jiang,
Xianming~Liu

\IEEEcompsocitemizethanks{\IEEEcompsocthanksitem Jialei Xu, Junjun Jiang, and Xianming Liu are with the School of Computer Science and Technology, Harbin Institute of Technology, Harbin, 150001, China, E-mail: xujialei@stu.hit.edu.cn, \{csxm, jiangjunjun\}@hit.edu.cn.
\IEEEcompsocthanksitem Wei Yin is with the SZ DJI Technology Co., Ltd. E-mail: yvanwy@outlook.com
\IEEEcompsocthanksitem Dong Gong is with The University of New South Wales. E-mail: edgong01@gmail.com.
}
\thanks{(Corresponding author: Xianming Liu)}
}


\maketitle
\begin{abstract}

Depth estimation is a critical technology in autonomous driving, and multi-camera systems are often used to achieve a $360\degree$ perception.
These $360\degree$ camera sets often have limited or low-quality overlap regions, making multi-view stereo methods infeasible for the entire image. Alternatively, monocular methods may not produce consistent cross-view predictions. To address these issues, we propose the \textbf{S}tereo \textbf{G}uided \textbf{D}epth \textbf{E}stimation (SGDE) method, which enhances depth estimation of the full image by explicitly utilizing multi-view stereo results on the overlap. We suggest building virtual pinhole cameras to resolve the distortion problem of fisheye cameras and unify the processing for the two types of $360\degree$ cameras. For handling the varying noise on camera poses caused by unstable movement, the approach employs a self-calibration method to obtain highly accurate relative poses of the adjacent cameras with minor overlap. These enable the use of robust stereo methods to obtain high-quality depth prior in the overlap region. This prior serves not only as an additional input but also as pseudo-labels that enhance the accuracy of depth estimation methods and improve cross-view prediction consistency. The effectiveness of SGDE is evaluated on one fisheye camera dataset, Synthetic Urban, and two pinhole camera datasets, DDAD and nuScenes. Our experiments demonstrate that SGDE is effective for both supervised and self-supervised depth estimation, and highlight the potential of our method for advancing downstream autonomous driving technologies, such as 3D object detection and occupancy prediction.
\end{abstract}

\section{Introduction}

\begin{figure}[t]
\center
\includegraphics[width=7.5cm,height=10cm]{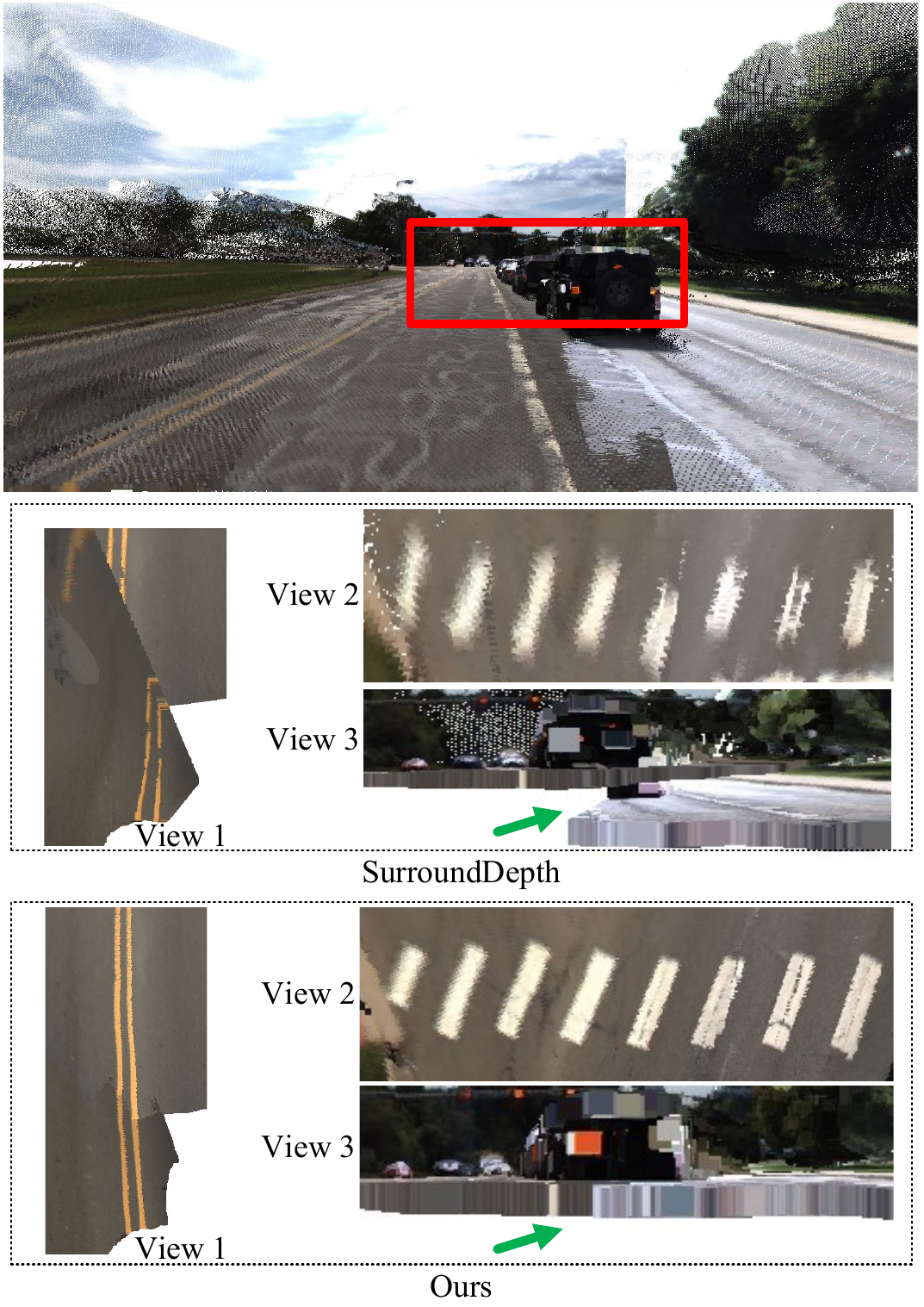}

\caption{\textbf{Multi-view point clouds from predicted depth.}
SGDE (ours) produces much more consistent point clouds across views.
View 3 shows the result of the reconstruction of the ground, our result of different views is on the same plane, while SOTA method (SurroundDepth~\cite{wei2022surrounddepth}) is not.
}
\label{fig:fist_image}

\end{figure}

%
\label{sec:intro}
Depth estimation is a fundamental technology for 3D scene perception in various computer vision applications, such as robotics and autonomous driving \cite{godard2019digging,ling2016high,guizilini20203d}. Benefiting from the advent of deep neural networks (DNNs), many deep learning based methods have been proposed to estimate the depth from monocular cues \cite{eigen2014depth,fu2018deep,Yin_2019_ICCV} or/and multi-view stereo cues \cite{scharstein2002taxonomy,hirschmuller2007stereo,teed2020raft}. In recent real-world applications, intelligent agents (\textit{e.g.}, autonomous driving cars) are equipped with \360 multi-camera sets including pinhole cameras \cite{caesar2020nuscenes,guizilini20203d} or fisheye cameras \cite{won2019sweepnet}. The \360 multi-camera sets bring a more reliable \360 field of view perception of the environment and also bring new challenges on how to properly and fully utilize them in depth estimation.

\par
\begin{figure}[t]
\flushleft

\centering
\subfloat{
\includegraphics[width=8.6cm]{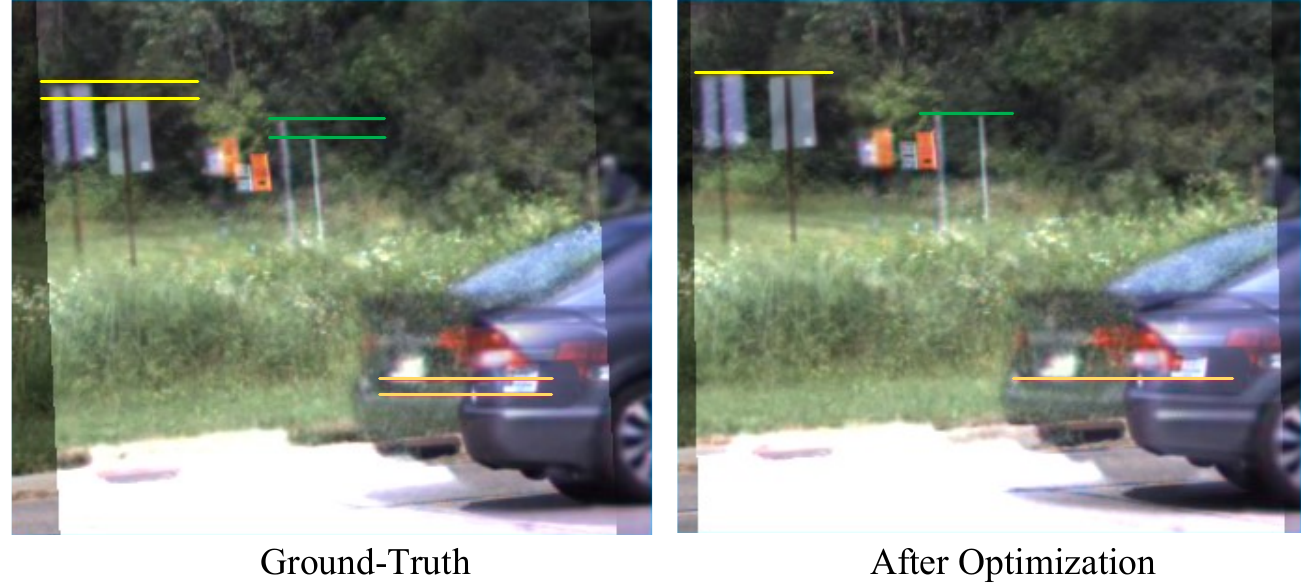}
}

\caption{
{ Overlaid stereo rectified images using the relative pose between adjacent two cameras in DDAD dataset~\cite{guizilini20203d}. 
The rectified images calculated according to the ``ground-truth'' pose provided by the dataset have a large error in the vertical direction, which indicates the inaccuracy of the dataset-provided parameters.
}
}

\label{fig:wrong pose}
\end{figure}

The \360 multi-view camera sets can provide powerful geometric cues for reliable 3D modeling. Instead of only using them as training supervision \cite{guizilini2022full,wei2022surrounddepth,xu2022multi}, the strong stereo information should be properly used in inference/testing to directly guide the depth estimation for robustness in the real world. Many multi-view stereo depth estimation methods \cite{yang2020cost,gu2020cascade} may pave the way for that. But they are non-ideal and inefficient for \360 depth estimation in the realistic applications (\textit{e.g.}, autonomous driving and robotics) that require immediate and efficient depth prediction for online response in the system. 
Although \360 multi-view camera sets can provide full coverage of the scene, the overlap region of the adjacent cameras is small (for surrounding pinhole cameras) \cite{guizilini20203d,caesar2020nuscenes}, or the overlapped regions are with low-quality/sparse signals (for surrounding fisheye cameras) \cite{won2019sweepnet}. On the other hand, in realistic scenarios, 
%
%
owing to the unstable mechanical structures during cars' movement, 
the pre-calibrated camera extrinsic parameters (i.e. cameras' translation and rotation to the car's ego system) may fail during, as shown in  Fig. \ref{fig:wrong pose}. 

\par
Existing \360 depth estimation methods \cite{guizilini20203d,wei2022surrounddepth,won2019omnimvs,won2020end} mainly focus on implicitly using the 
texture consistency information in the overlapping areas. Many methods \cite{guizilini2022full,wei2022surrounddepth,xu2022multi} only use \360 geometric cues to provide supervision in training (\textit{e.g.}, self-supervised depth estimation methods) and cannot guarantee the multi-view consistency in overlapping areas during inference, which cannot fully use the geometric guidance in the \360 images. 
On the other hand, they mainly focus on pinhole cameras \cite{guizilini20203d,caesar2020nuscenes}, and the studies on \360 fisheye cameras \cite{won2019sweepnet,won2019omnimvs,won2020end} overly rely on synthetic datasets. For both of them, existing works focus on the ideal scenarios and may ignore or overlook the characteristics of the realistic \360 images, \textit{e.g.}, low-quality overlapping area and the pose measurement noise, as discussed above.


\par
In this paper, we address the above issues by proposing a general pipeline named Stereo Guided Depth Estimation (SGDE) for \360 depth estimation, which \emph{explicitly} exploits the geometric cues in the overlapping regions in both inference and training. 
In our approach, we propose to use classical stereo-matching methods \cite{lipson2021raft,hirschmuller2005accurate} to infer depth in the overlapping regions as a \emph{depth prior} for more accurate depth estimation with DNNs, which can be very efficient and thus practical for the real-world application scenarios of the \360 depth estimation, comparing to previous MVS methods \cite{gu2020cascade,yang2020cost}. 
By triangulating the depth prior on the overlap, we can obtain dense depth maps on the overlapping regions as prior and use them as input guidance (usable in inference) for DNNs or/and pseudo ground truth to provide supervision (in training). The explicit stereo guidance enables more powerful guidance and robust predictions than other implicit techniques \cite{guizilini2022full,wei2022surrounddepth,xu2022multi}. 
Specifically, to unify the process for different types of cameras, we propose to build virtual pinhole cameras for the fisheye cameras via camera model transformation. 
For the \360 fisheye cameras with significant overlap but serious image quality problems (i.e., uneven pixel density and distortion), the accuracy of the high-quality unsupervised depth prior exceeds multiple supervised methods.
For the \360 pinhole cameras with small overlaps, our method can significantly improve the accuracy, particularly mitigating depth consistency issues on the overlap.
Moreover, to handle the varying measurement noise/error on camera poses, we propose to dynamically optimize and update the camera pose in the proposed framework, which guarantees the effectiveness of stereo rectification in realistic applications (as discussed in Sec. \ref{sec:pose_optim}). We summarize our contributions in the following:

%

\begin{itemize}
\item We target the \360 view depth estimation problem in autonomous driving scenarios, and propose a general pipeline to solve the \360 fisheye and pinhole camera sets with explicit stereo guidance, called Stereo Guided Depth Estimation (SGDE).

\item We propose to build virtual pinhole cameras to solve the distortion problem of uneven pixel density and serious distortion of fisheye cameras in the geometry method, which unifies the process of the two types of \360 camera sets. 

\item To handle the varying measurement noise/caused by unstable movement in realistic scenarios, we introduce the geometric loop constraint to calibrate camera poses by surrounding cameras with small overlap, which achieves higher accuracy than the ground truth and ensures the effectiveness of SGDE. 

\item We present experimental results on Synthetic Urban ~\cite{won2019sweepnet} and two popular real multi-camera datasets DDAD~\cite{guizilini20203d} and nuScenes~\cite{caesar2020nuscenes} showing that SGDE can effectively improve the performance of SOTA models in both supervised and self-supervised scenarios. We highlight the potential of our method for downstream autonomous driving technologies, such as object detection and occupancy prediction.

\end{itemize}

\section{Related Work}

\noindent\textbf{Monocular Depth Estimation}.
Monocular depth estimation aims at predicting the depth map from a single image. 
Eigen~\textit{et al.} ~\cite{eigen2014depth} propose the pioneered approach which inspiresd many follow-up works. The CNN-based method employse two deep network stacks: one that makes a coarse global prediction based on the entire image, and another that refines this prediction locally. Subsequent works propose to directly regress the continuous depth map from the aggregation of the information in an image~\cite{liu2015learning,lee2019big,song2021monocular,alhashim2018high}, and continuously improve the accuracy. ~\cite{fu2018deep,bhat2021adabins} formulate depth prediction into classification,
and predict the depth class of each pixel to get the depth map. 
Besides the depth ground-truth, 
Yin~\textit{et al.} ~\cite{Yin_2019_ICCV} show the geometric information in the 3D space can also help the model to predict the depth. 
Huynh~\textit{et al.} ~\cite{huynh2020guiding} propose guiding depth estimation to favor planar structures that are ubiquitous especially in indoor environments. 
Due to the heavy cost of obtaining high quality ground-truth from depth-sensors, self-supervised depth prediction has attracted much attention in recent years. 
Zhou~\textit{et al.}~\cite{zhou2017unsupervised} propose a fundamental framework consisting of a depth network and a pose network. The depth network estimated the depth of frame $I_t$ supervised by consecutive ones\{$I_{t-1}$,$I_{t}$ $I_{t+1}$\}, and the pose network estimated the pose between the $I_{t}$ and $I_{t {\pm} 1}$. Follow-up works~\cite{godard2019digging,gonzalezbello2020forget,guizilini20203d,guizilini2020semantically} design better model architectures and loss functions in this direction. 


\noindent\textbf{Multi-Camera Depth Estimation}.
Stereo matching has been an active research area for decades~\cite{scharstein2002taxonomy,hamzah2016literature}. Traditional methods~\cite{hosni2012fast,hirschmuller2007stereo,scharstein2002taxonomy} focuse on looking for local matching points and global optimization to exploit spatial context. Recently, deep learning has brought big improvements to stereo matching. The end-to-end networks~\cite{mayer2016large, ranjan2017optical} learn stereo matching by assembling multiple steps together, yielding more accurate results. Kendall~\textit{et al.}~\cite{kendall2017end} leverage the geometric knowledge to form a cost volume using deep feature representations, which allow the network to run per-pixel feature matching. 
Lipson~\textit{et al.} ~\cite{lipson2021raft} introduce RAFT-Stereo, a new deep architecture for rectified stereo based on the optical flow network RAFT~\cite{teed2020raft}. With multi-level convolutional GRUs and the network trained on synthetic data, it shows high accuracy.

As the demand for $360\degree$ surround perception in autonomous driving continues to rise, there is an increased focus on depth reconstruction by multi-camera in surrounding views.
In Synthetic Urabn dataset~\cite{won2019sweepnet}, the camera rig is virtually implemented with four fisheye cameras, and ~\cite{won2019sweepnet, won2019omnimvs,won2020end} rely on the large areas of overlap and perfect depth ground-truth to predict panorama depth. However, the high-quality ground-truth is hard to obtain in real data.
In surrounding pinhole cameras setups, there is only a small overlap between adjacent cameras such as the DDAD dataset~\cite{guizilini20203d} and nuScenes dataset~\cite{caesar2020nuscenes}, which opens up a new direction for multi-camera depth prediction. Guizilini~\textit{et al.}~\cite{guizilini2022full} first propose the self-supervised learning of scale-aware network based on monodepth in the multi-camera setting. SurroundDepth~\cite{wei2022surrounddepth} employ a joint network to process all the surrounding views and proposed a cross-view transformer to effectively fuse the information from multiple views. 
Xu~\textit{et al.}~\cite{xu2022multi} formulate the depth estimation as a weighted combination of depth basis.
However, these methods only implicitly learn the information in the overlapping area, which is less effective.
We propose to explicitly infer the depth prior in the overlap by stereo method before training models, which can make maximum use of the geometric information of the overlap.

\noindent\textbf{Multi-Camera Calibration}.
Camera calibration has important research and application values in computer vision. 
There are two main approaches to the calibration problem: 1) \emph{Chart-based calibration} assumes that an object with precisely known geometry is presented in all images, and computes the camera parameters consistent with a set of correspondences between the features defining the chart and their observed image projections. 2) Structure from Motion (SfM)~\cite{hartley2003multiple}, which estimates both the scene shape and the camera parameters. In SfM, the intrinsic parameters are either known a prior or recovered a posteriori through auto-calibration. Typically, camera parameters and scene points are computed by non-linear optimization during bundle adjustment~\cite{triggs1999bundle}. The selection of feature correspondences is a key step of both approaches.
SIFT~\cite{ng2003sift} and ORB\cite{rublee2011orb} are arguably the most successful hand-crafted local features and are widely adopted in many 3D 
vision tasks. 
In the surrounding pinhole camera case, insufficient overlap area between cameras can negatively impact the performance of camera calculation in existing methods.  To address this issue, we propose to use the relative pose of adjacent cameras to form a loop in geometry, which effectively improves the accuracy and robustness of the pose optimization.

\begin{figure*}[t!]
\centering

\includegraphics[width=17.5cm]{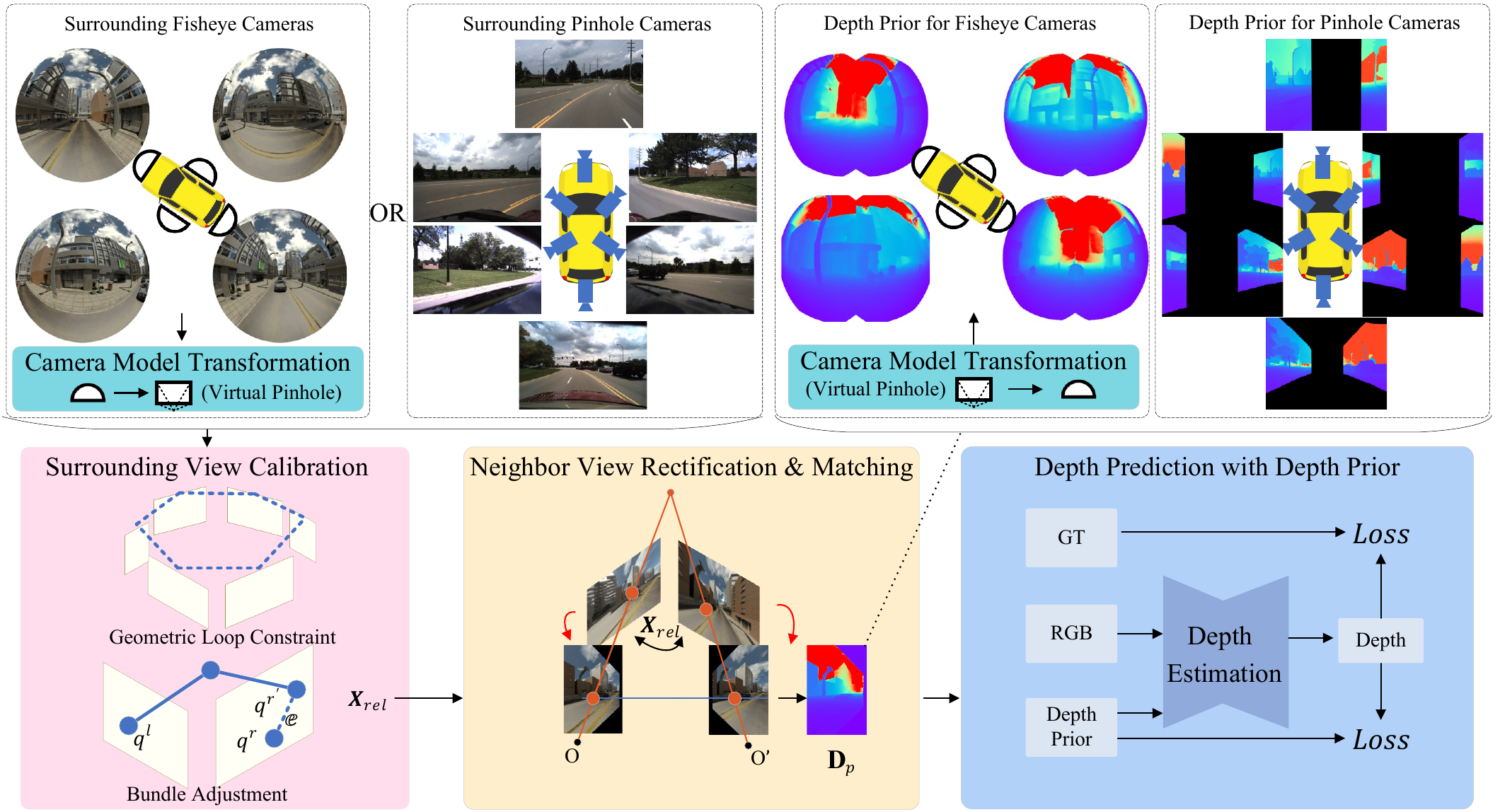}

\caption{\textbf{The pipeline of Stereo Guided Depth Estimation (SGDE).} 
For the \360 camera set, we first propose the geometry loop constraint to optimize the surrounding camera poses, which ensures the effectiveness of the subsequent geometry scheme. Then, we use the mature stereo rectification and matching algorithms to obtain the depth prior of the overlapping regions. Finally, depth prior is used as both an extra input and the supervisor signal to enhance the depth estimation network.
}
\label{fig:pipline}

\end{figure*}

\section{The Proposed Method}
Multiple cameras, such as pinhole cameras or fisheye cameras, are usually equipped in real-world applications (\textit{e.g.}, autonomous cars) to obtain surround $360\degree$ views. We aim to solve this $360\degree$ view depth estimation task by proposing a stereo guided pipeline, as shown in Fig.~\ref{fig:pipline}. We observe that all adjacent cameras have some overlaps, 
which can provide confident multi-view observation of the scene at the same time. 
We propose to 
obtain the depth prior of the overlapping areas and use it explicitly to enhance the performance of depth prediction networks. 
%
This motivates us to explicitly use the stereo-matching methods to obtain the depth in overlapping regions and  then use them to supervise training or as input. However, in autonomous driving scenarios, the adjacent cameras are not normal stereo cameras, such as completely different focal lengths, noisy camera extrinsic parameters, and two fisheye cameras. We propose several modules to solve them, including camera model transformation, surrounding view calibration, and neighbor view rectification and matching.

\subsection{Depth Prior Construction}\label{sec:stereo_rectify}

\subsubsection{Camera Model Transformation}
\label{sec:camera model transformation}
We first propose to build virtual pinhole cameras to solve the problem of uneven pixel density and serious distortion of surrounding fisheye cameras in the stereo-matching methods, which can unify the process for pinhole and fisheye cameras. 
Specifically, the field of view in one fisheye camera is evenly divided into left and right parts corresponding to two virtual pinhole cameras. Using one virtual camera as an example, we manually set the intrinsic parameters $\mathbf{K}_{pin}$ and extrinsic parameters 
$\mathbf{X}_{pin} = \begin{psmallmatrix}\mathbf{{R}}_{pin} & \mathbf{{t}}_{pin}\\ \mathbf{0} & 1\end{psmallmatrix}$ for the virtual pinhole camera. The fisheye image is transformed to the pinhole image with the following warping. In the fisheye image, we suppose a point located at $\mathbf{p}$ = $(\Phi, \Theta)$ in the 2D spherical coordinates, and is transformed to the spherical projection of the space point $\mathbf{\hat{p}}$:
\begin{equation}
\mathbf{\hat{p}} =  [\text{sin}(\Phi) \text{cos}(\Theta), \text{sin}(\Phi) \text{sin}(\Theta), \text{cos}(\Phi)]^\mathrm{T},
\label{eq:fisheye to pinhole}
\end{equation}
where $\hat{\mathbf{p}}$ is the unit directional vector. Then, $\hat{\mathbf{p}}$ is projected to the point $\mathbf{p}_{pin} = (u, v)$ in image coordinates of the pinhole camera as following: $\mathbf{p}_{{pin}}$  = 
$ \mathbf{K}_{pin} {\mathbf{R}_{pin}}^{-1} ( \mathbf{R}_f \mathbf{\hat{p}} + \mathbf{t}_f - \mathbf{t}_{pin})$,
where $\mathbf{X}_f = \begin{psmallmatrix}\mathbf{{R}}_f & \mathbf{{t}}_f\\ \mathbf{0} & 1\end{psmallmatrix}$
 is the extrinsic parameters of fisheye cameras. 
We also perform linear interpolation sampling to compute the virtual pinhole image.
Here, we obtain the pinhole images of the \360 views from fisheye cameras and corresponding calibrations.


\subsubsection{Neighbor View Rectification and Matching}
We can then derive the relative pose between each pair of adjacent cameras in the same frame from datasets or the module of camera model transformation (Sec.~\ref{sec:camera model transformation}), and use standard stereo methods to reconstruct a reliable depth prior on the overlap. 
The overall process of depth prior construction includes three steps: 
1) \emph{Stereo rectification}, which projects overlapping images into a common image plane and decides the overlapping area. 2) \emph{Stereo matching}, which computes the depth of the overlap region based on known camera intrinsic and extrinsic. 3) \emph{Depth back-projection}, which back-projects the depth from the common image plane to the origin camera plane. We call it depth prior $\mathbf{D}_p$. 
We use an off-the-shelf stereo matching algorithm, e.g., SGBM~\cite{hirschmuller2005accurate} or Raft-Stereo~\cite{lipson2021raft}, to obtain $\mathbf{D}_p$. Note that we do not re-train Raft-Stereo, but directly use the model trained on the Middlebury dataset~\cite{scharstein2014high} for depth reconstruction. SGBM is a non-parametric method.

\begin{figure}[th!]
\flushleft

\centering
\includegraphics[width=\linewidth]{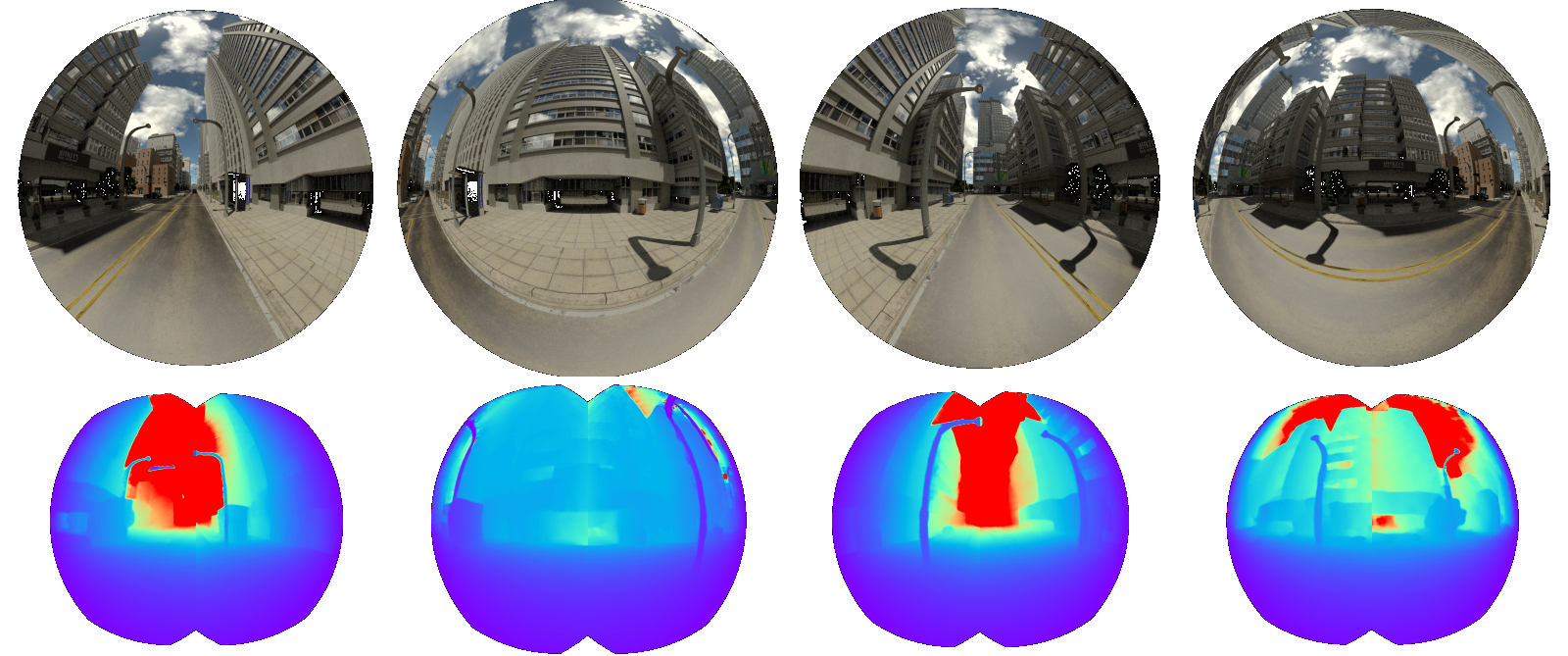}
\\
\includegraphics[width=\linewidth]{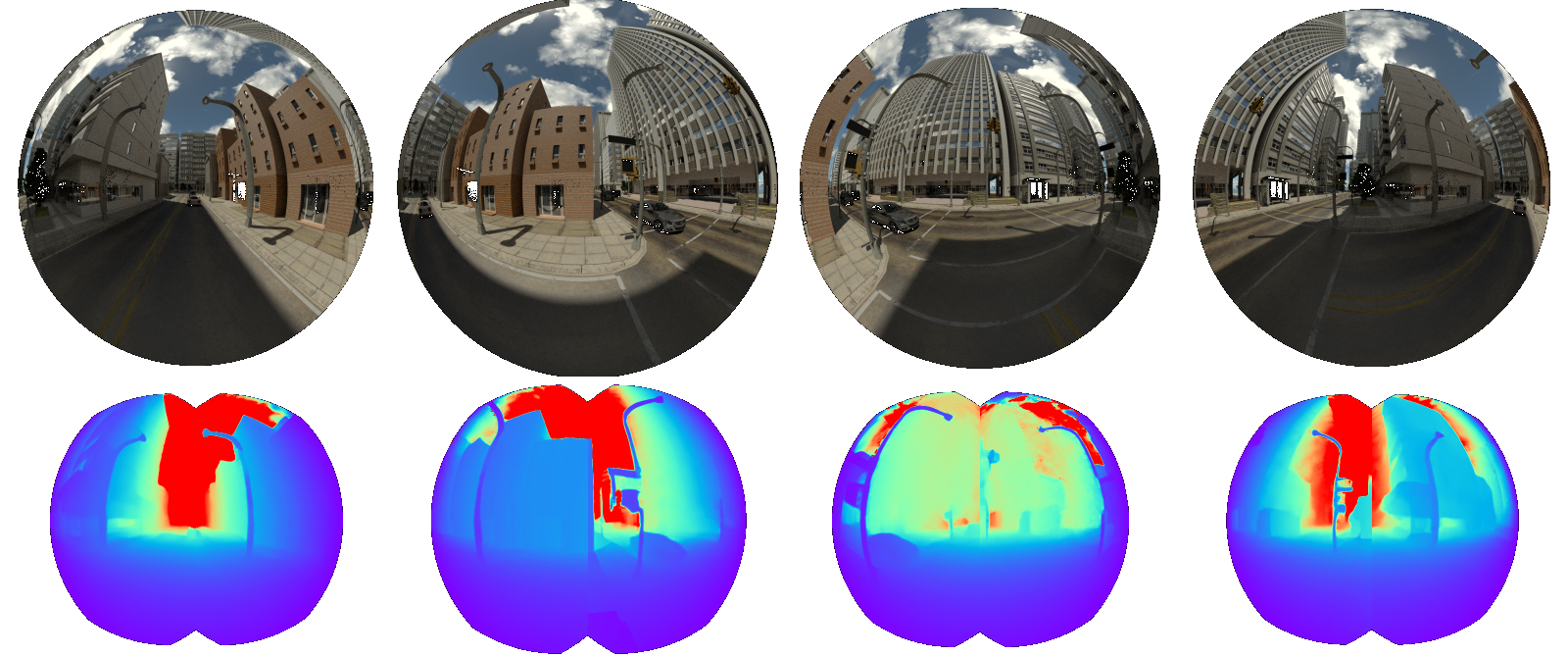}
\\
\includegraphics[width=\linewidth]{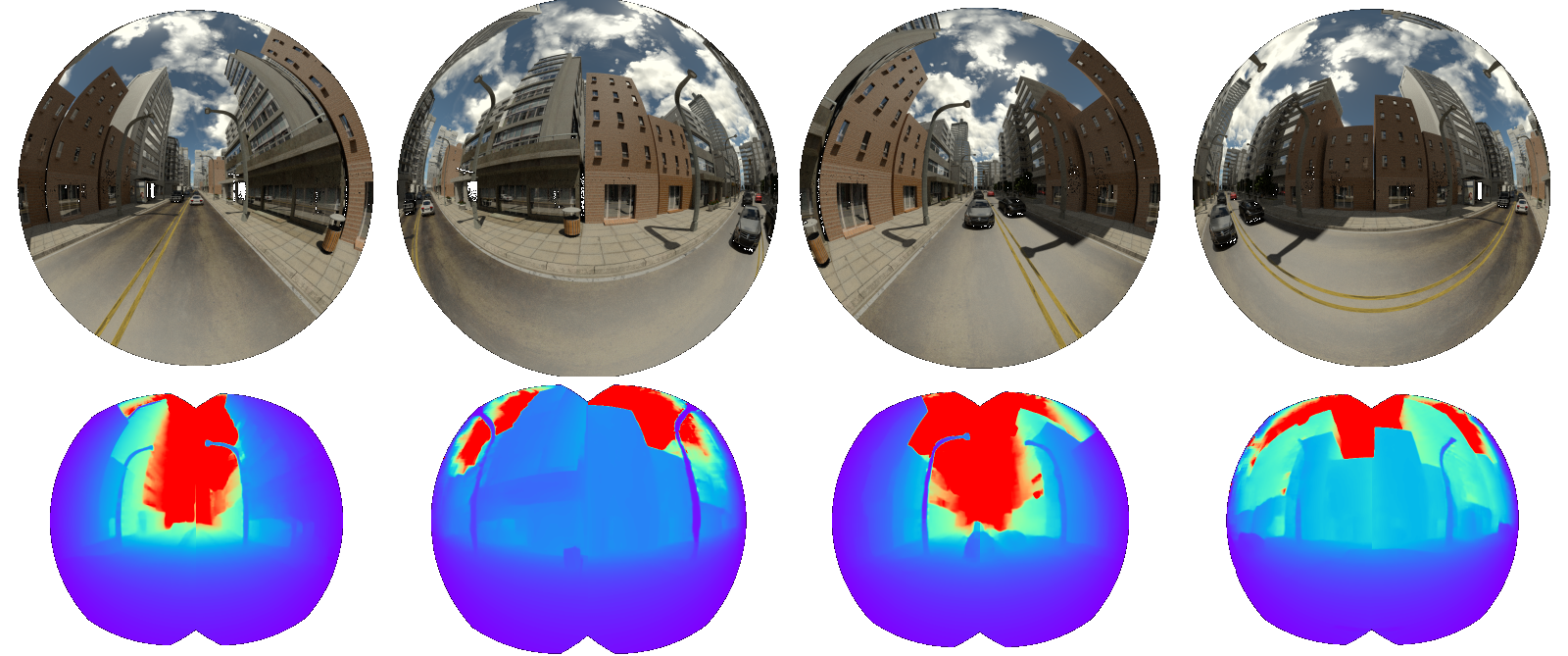}
\includegraphics[width=\linewidth]{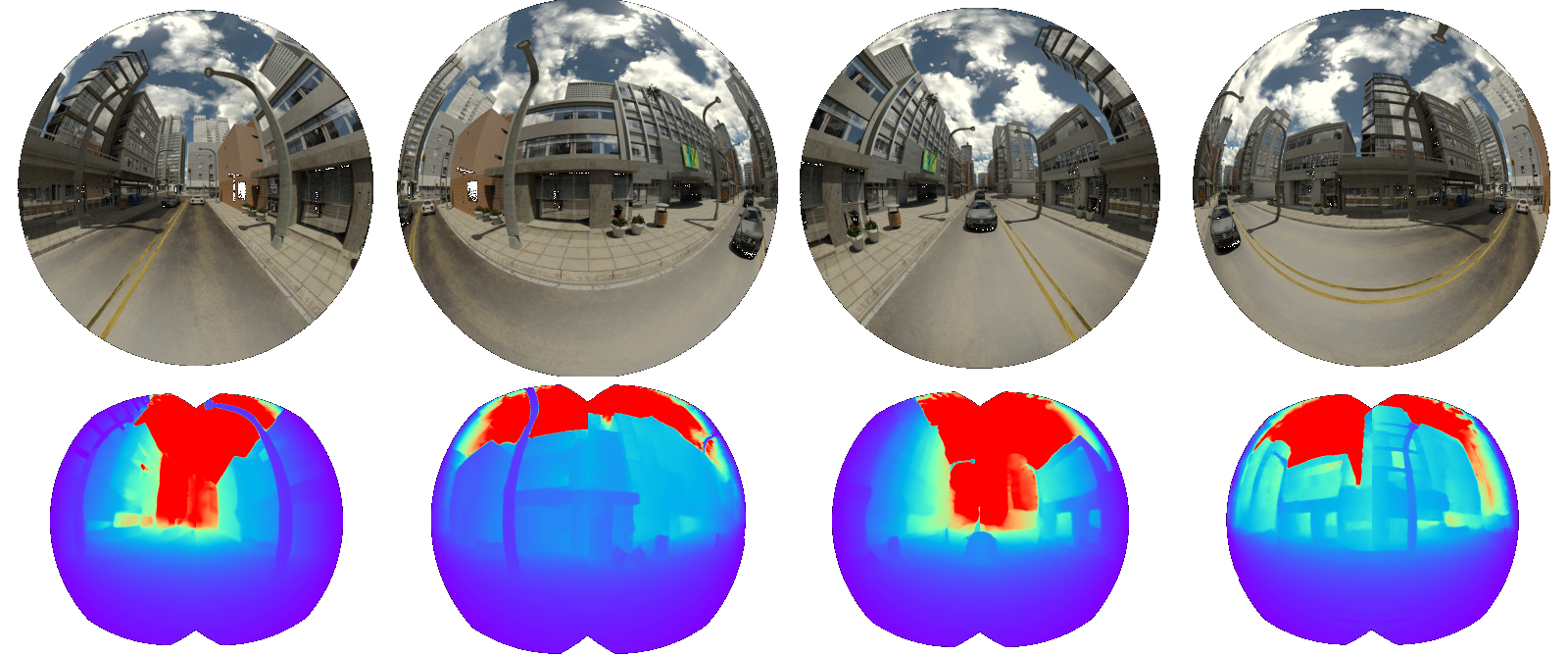}
\caption{
\textbf{Depth prior in Synthetic Urban~\cite{won2019sweepnet} dataset, which is computed by RAFT-Stereo~\cite{lipson2021raft}.}
}
 
\label{fig:fisheye}
\end{figure}


\subsection{Surrounding View Calibration}\label{sec:pose_optim}
\label{sec:Calibration Optimization}
Though the process of depth prior construction seems straightforward, it heavily relies on the accuracy of camera calibrations.
The relative poses between cameras provided in common autonomous driving datasets are very noisy or even completely fail. This is because the vibration during car driving is likely to change the location or orientation of the camera. As shown in Fig.~\ref{fig:wrong pose}, we overlay the rectified images using the relative poses provided in the dataset~\cite{guizilini20203d}. 
The same object is not on the horizontal epipolar line, which  indicates the inaccuracy of the dataset-provided calibration parameters.
Such noisy calibrations will cause failure in stereo matching and depth calculation. 
To test this point, we evaluate the depth prior computed by the calibration parameters from datasets. As shown in Tab.~\ref{tab:overlap result}, using the ground-truth camera calibrations (provided in the datasets) results in unacceptable depth reconstructions in both experimented datasets. Therefore, we argue that the ground-truth camera parameters, specifically, the camera poses provided in common autonomous driving datasets are not accurate enough for the depth prior construction. 

To address this problem, we need to optimize ``ground-truth'' camera poses between adjacent cameras.
The process of high-precision stereo camera calibration is similar. By extracting correspondences in the overlap region between cameras and accumulating them over multiple frames, the relative pose can be computed by the method of epipolar geometry. However, we find this strategy to be ineffective in DDAD~\cite{guizilini20203d} and nuScenes~\cite{caesar2020nuscenes}, as the overlap region is too small. Insufficient overlap results in the failure of two-view geometry frequently as shown in Tab.~\ref{tab:overlap result}. Even with the SOTA stereo camera calibration method HOM~\cite{ling2016high}, the error of the depth prior computed by the optimized pose is still significant. Hence, we propose 
to establish the geometric loop constraint (GLC) in \360 views to calibrate the surrounding cameras' pose synchronously. We use the relative poses of the \360 views cameras to form a loop, which has a stronger restraint ability in the optimization process than the ordinary stereo calibration schemes.

Specifically, we optimize the poses for all cameras within a video sequence. The overall process for our calibration method contains 3 steps: 1) \emph{Correspondence extraction}, which finds correspondences between all images within a small number of frames. 2) \emph{Pose parametrization}: we apply a temporal consistency assumption, and assume that the relative pose between cameras remains consistent in a short period of time. 3) \emph{Bundle adjustment}: 
the pose of all cameras is optimized simultaneously, making it possible to create a geometric loop constraint on the \360 views,  and compute the final poses 
of all cameras based on the extracted correspondences and pose parametrization.

Given a video sequence with $T$ frames and $M$ cameras, we denote $\mathbf{I}_{t}^m$ as the image from camera $m$ in frame $t$.
We adopt LoFTR~\cite{sun2021loftr} to extract candidate matching from all possible $T \times T \times M$ pairs of images from adjacent cameras, and then filter outlier matching and get the final correspondences $\{\mathbf{q}^{m_1}_{t_1},\mathbf{q}^{m_2}_{t_2}\}$ for each image pair \{$\mathbf{I}^{m_1}_{t_1}$,$\mathbf{I}^{m_2}_{t_2}$\} using Graph-Cut RANSAC~\cite{barath2018graph}.
Denote the forward-facing camera id $m = 1$, we first parametrize the pose $\mathbf{X}^1_t$ of the forward-facing camera in each frame $t$. For other cameras, we only optimize the relative pose $\mathbf{X}^{m}_{\text{rel}}$ from each camera to the forward-facing camera. The absolute pose $\mathbf{X}^{m}_{t}$ of camera $m > 1$ in frame $t$ can thus be computed by:
\begin{align}\label{eq:pose_param}
\mathbf{X}^{m}_t &=  \mathbf{X}^{m}_{rel} \mathbf{X}^1_t.
\end{align}
%
We optimize all surrounding cameras synchronously to establish geometry loop constraint, and use standard bundle adjustment~\cite{hartley2003multiple} to optimize the poses:
\begin{equation}
\underset{\{\mathbf{X}^{1}_{t}\}, \{\mathbf{X}^{m}_{rel}\}, \{\mathbf{\hat{q}}_i\}}{\text{minimize}} \sum_{m=1}^M \sum_{t=1}^T \sum_{i=1}^N | \mathbf{q}_{t,i}^m-{\pi(\mathbf{K}^{m}, \mathbf{X}^{m}_t, \mathbf{\hat{q}}_i)} |,
\end{equation}
where  $\mathbf{K}^m$ is the intrinsic matrix of camera $m$, 
$\mathbf{q}_{t,i}^m \in \mathbf{q}^m_{t}$ 
is the 2D pixel location of an observed 3D point $\mathbf{\hat{q}}_i$ in image $\mathbf{I}_t^m$, $N$ is the number of observed 3D point $\mathbf{\hat{q}}$, and $\pi(\cdot)$ is the projection function from a 3D point to a camera plane. Intuitively, the above equation jointly optimizes all the $T+M-1$ poses along with the 3D location of all correspondences.

\begin{figure*}[th!]
\flushleft

\centering
\subfloat{
\parbox[t]{2mm}{\rotatebox{90}{\small~~~~RGB}}
\includegraphics[width=2.9cm,height=1.74cm]{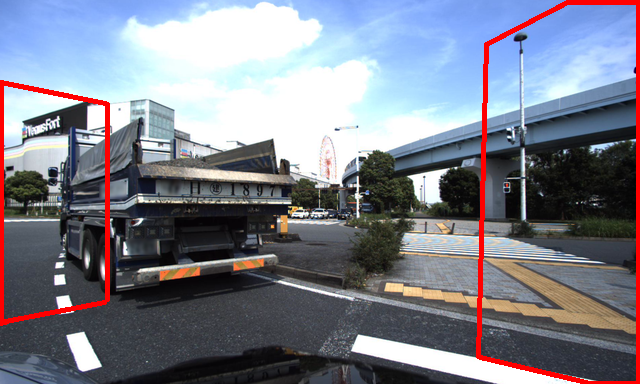}\hspace{-0.25em}
\includegraphics[width=2.9cm,height=1.74cm]{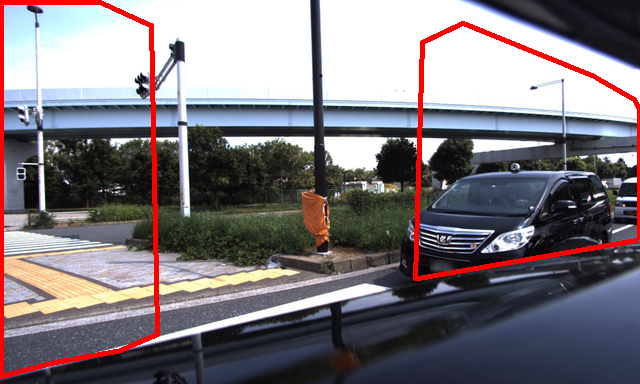}\hspace{-0.25em}
\includegraphics[width=2.9cm,height=1.74cm]{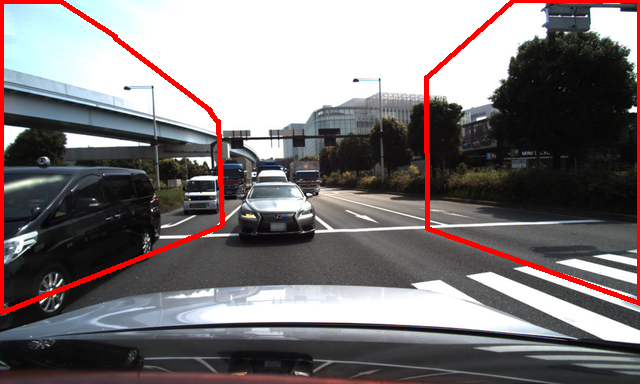}\hspace{-0.25em}
\includegraphics[width=2.9cm,height=1.74cm]{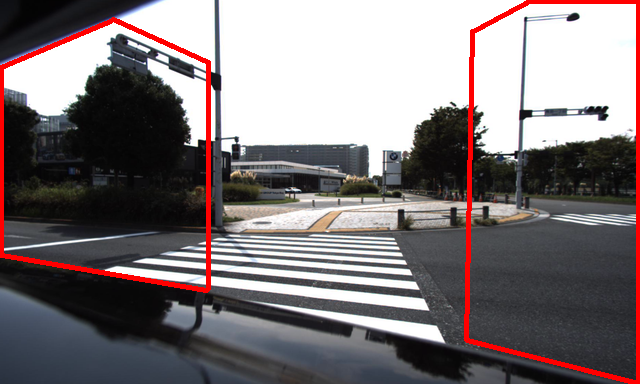}\hspace{-0.25em}
\includegraphics[width=2.9cm,height=1.74cm]{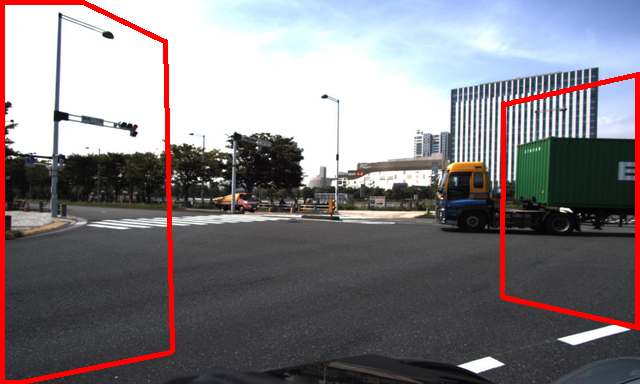}\hspace{-0.25em}
\includegraphics[width=2.9cm,height=1.74cm]{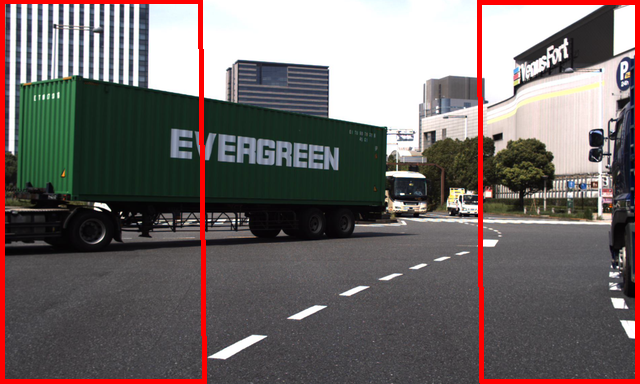}
}
\vspace{-0.8em}
\\
\subfloat{
\parbox[t]{2mm}{\rotatebox[origin=c,x=0mm,y=0.0mm]{90}{\small~~~Overlap}}
\includegraphics[width=2.9cm,height=1.74cm]{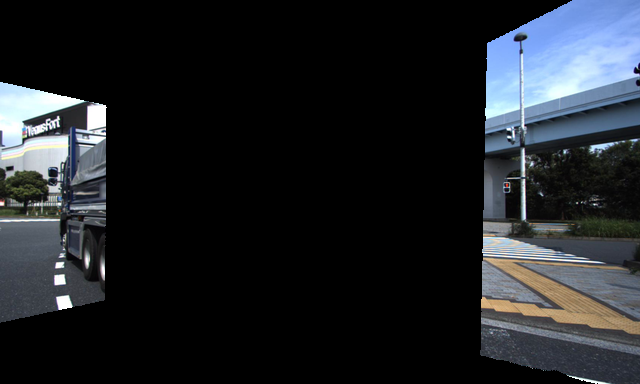}\hspace{-0.25em}
\includegraphics[width=2.9cm,height=1.74cm]{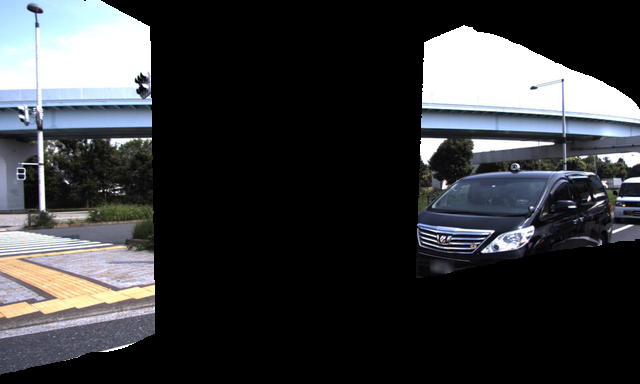}\hspace{-0.25em}
\includegraphics[width=2.9cm,height=1.74cm]{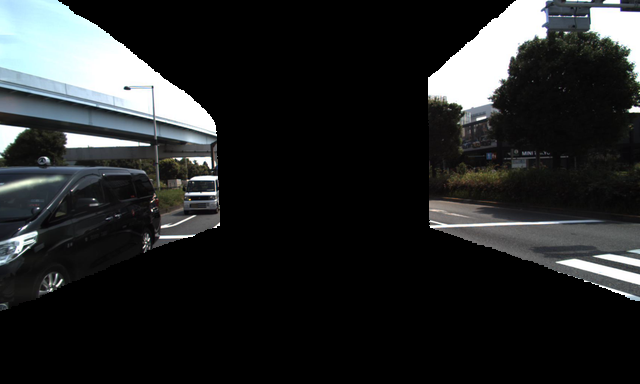}\hspace{-0.25em}
\includegraphics[width=2.9cm,height=1.74cm]{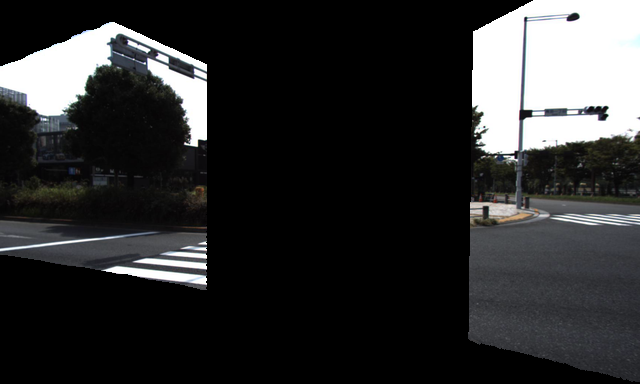}\hspace{-0.25em}
\includegraphics[width=2.9cm,height=1.74cm]{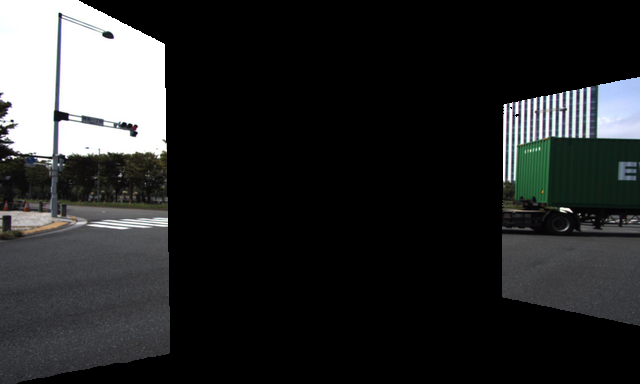}\hspace{-0.25em}
\includegraphics[width=2.9cm,height=1.74cm]{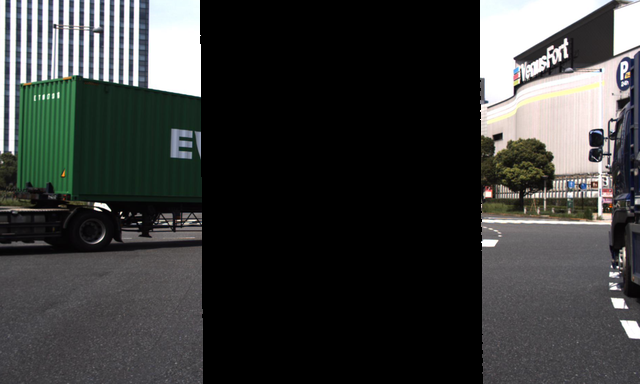}
}
\vspace{-0.8em}
\\
\subfloat{
\parbox[t]{2mm}{\rotatebox[origin=c,x=0mm,y=0.0mm]{90}{\small Depth Prior}}
\includegraphics[width=2.9cm,height=1.74cm]{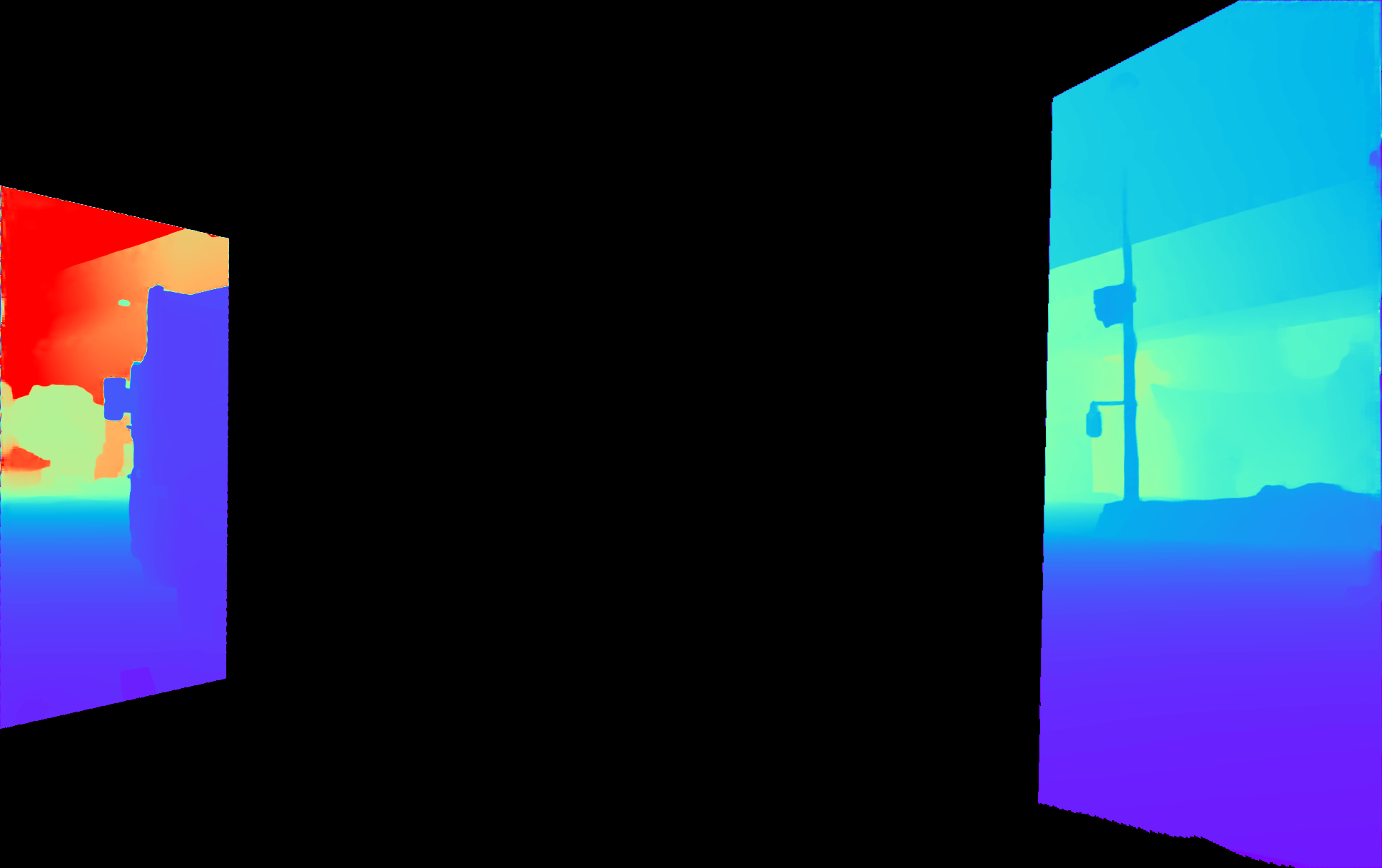}\hspace{-0.25em}
\includegraphics[width=2.9cm,height=1.74cm]{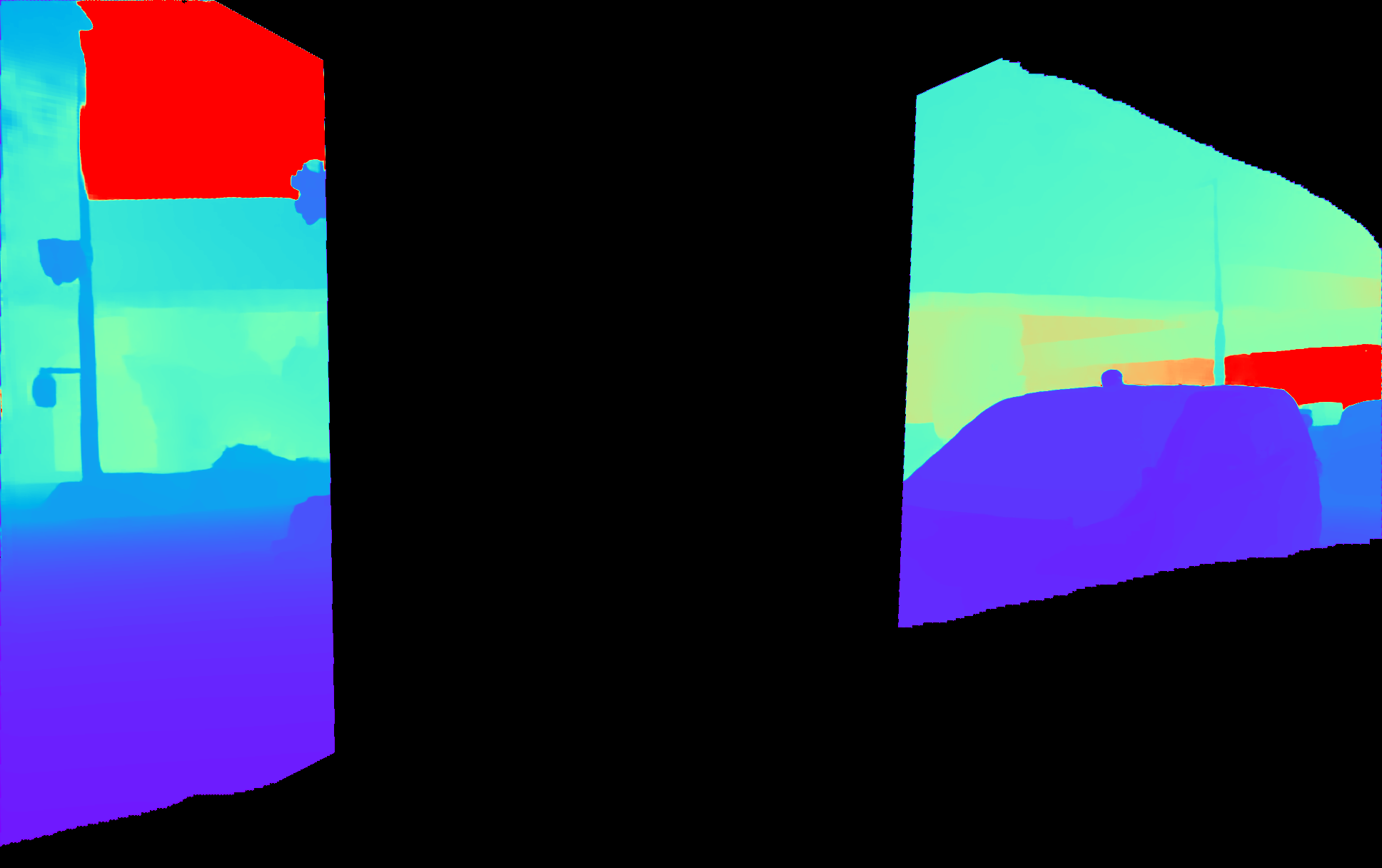}\hspace{-0.25em}
\includegraphics[width=2.9cm,height=1.74cm]{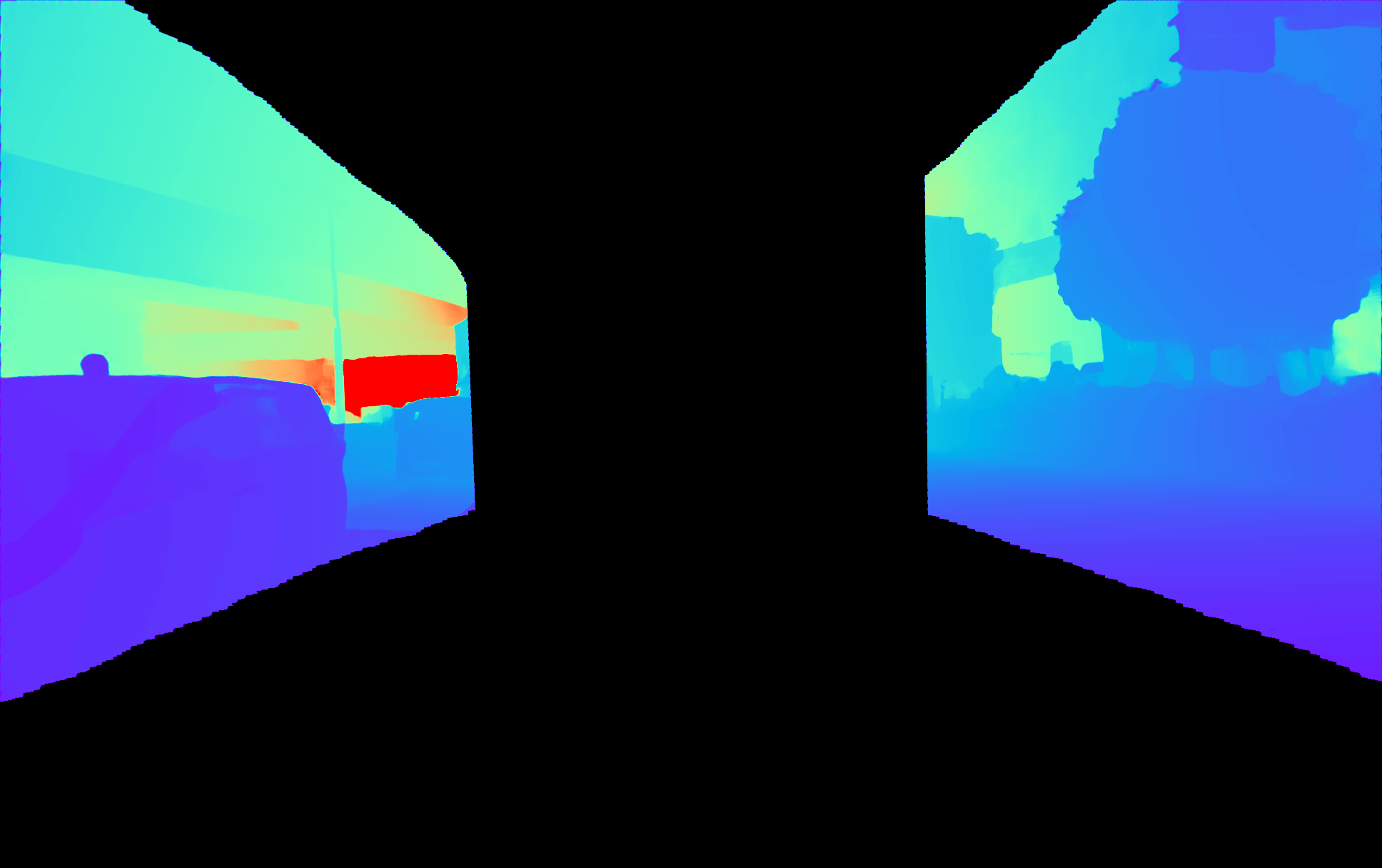}\hspace{-0.25em}
\includegraphics[width=2.9cm,height=1.74cm]{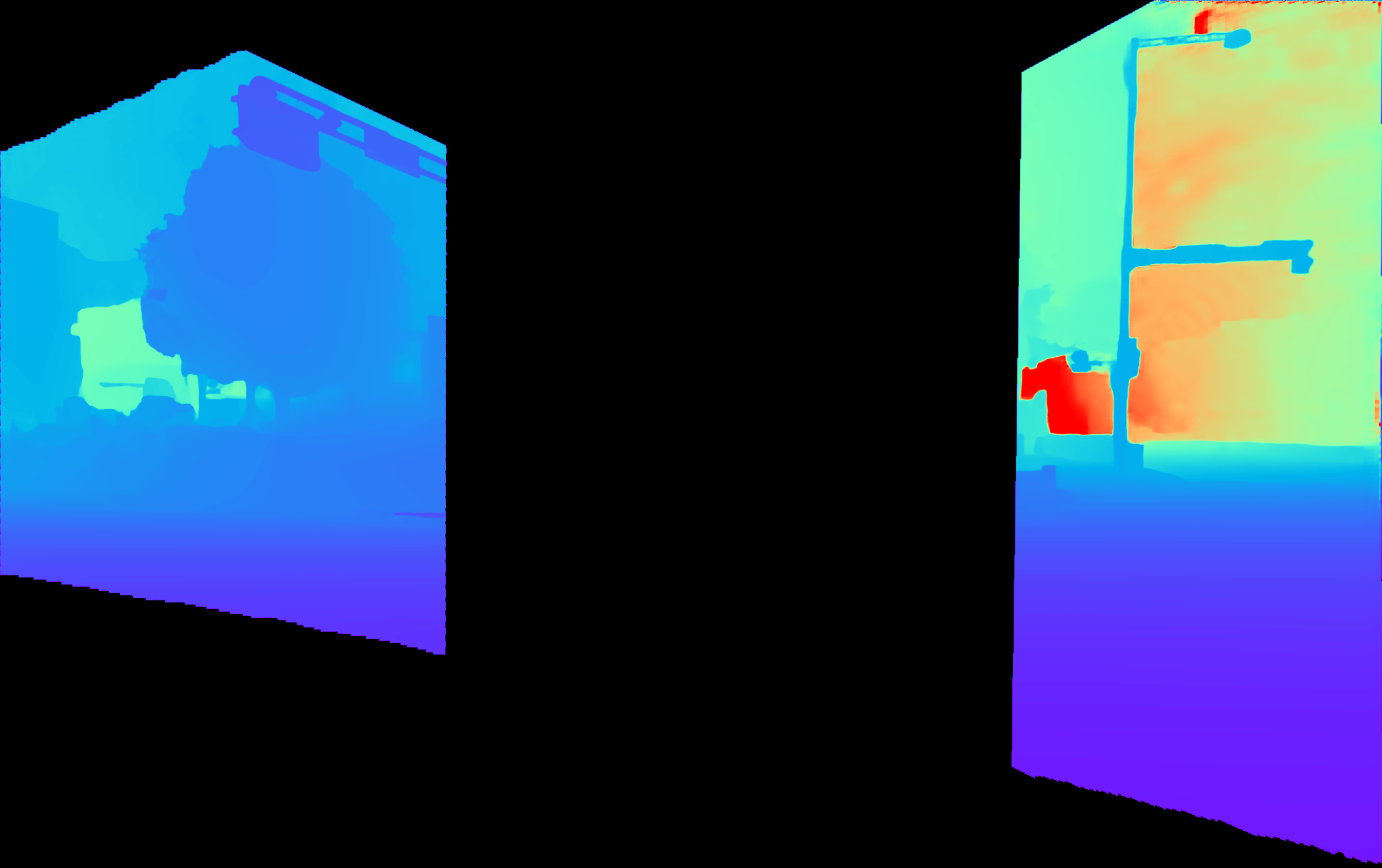}\hspace{-0.25em}
\includegraphics[width=2.9cm,height=1.74cm]{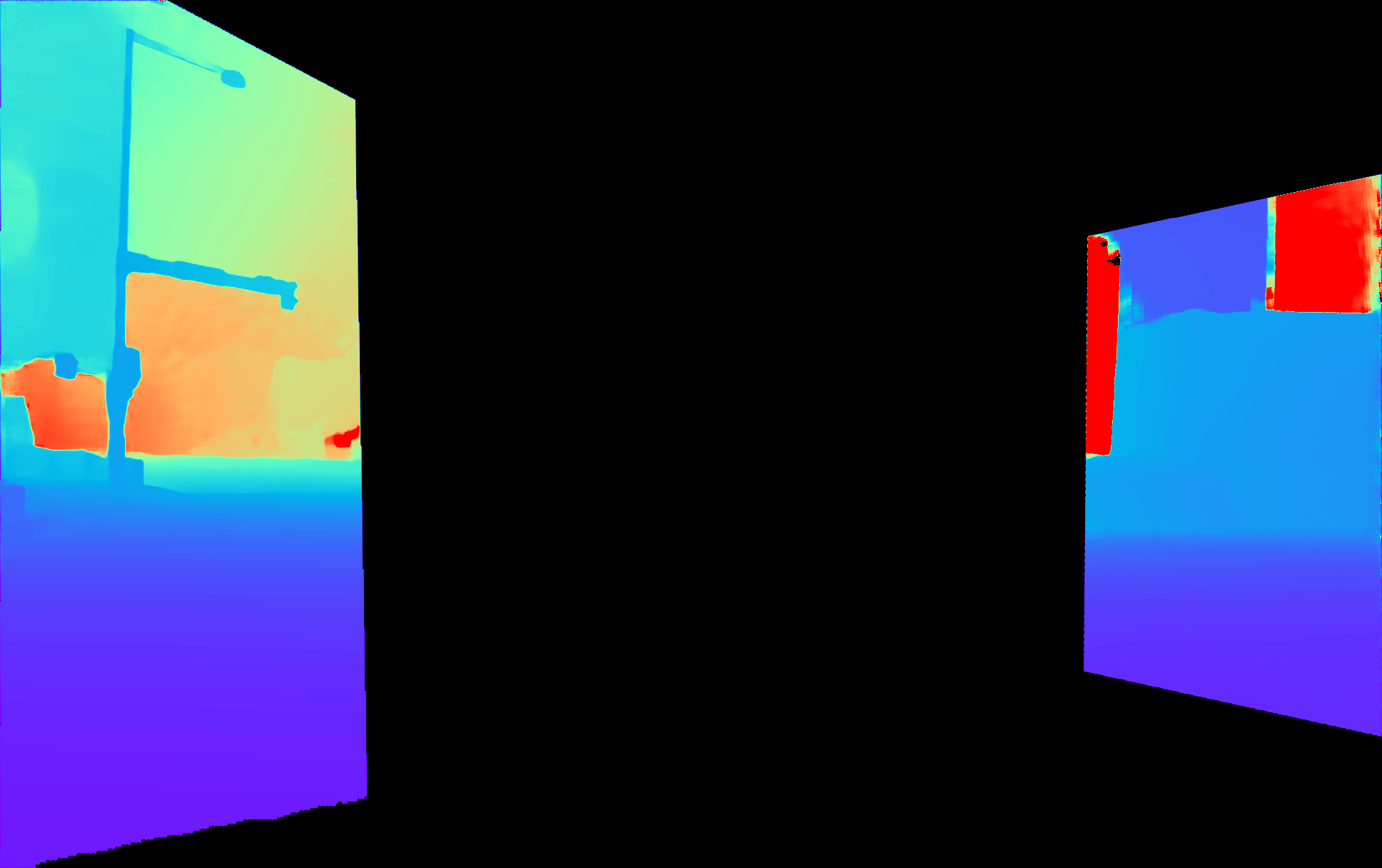}\hspace{-0.25em}
\includegraphics[width=2.9cm,height=1.74cm]{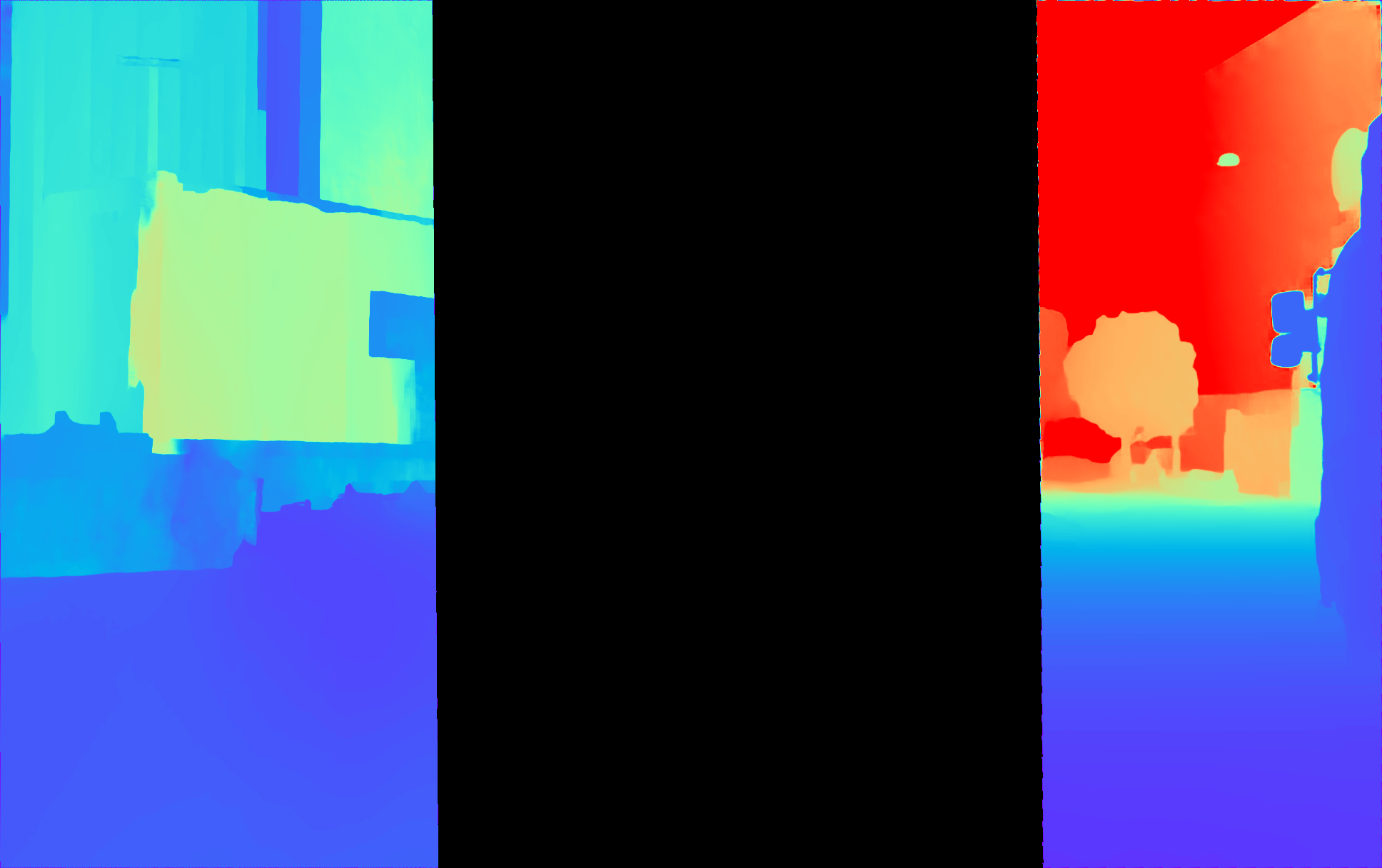}
}
\vspace{-0.8em}
\\
\subfloat{
\parbox[t]{2mm}{\rotatebox[origin=c,x=0mm,y=0.0mm]{90}{\small~~R18\cite{godard2019digging}}}
\includegraphics[width=2.9cm,height=1.74cm]{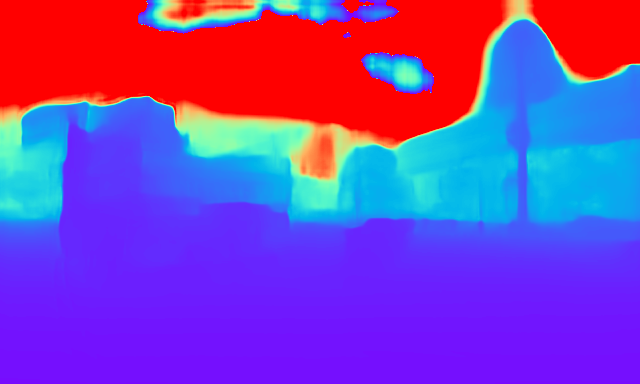}\hspace{-0.25em}
\includegraphics[width=2.9cm,height=1.74cm]{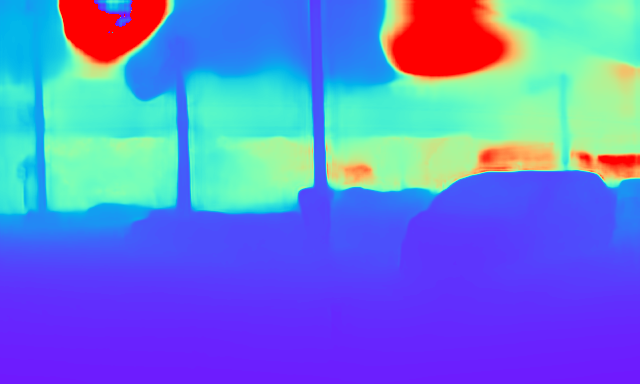}\hspace{-0.25em}
\includegraphics[width=2.9cm,height=1.74cm]{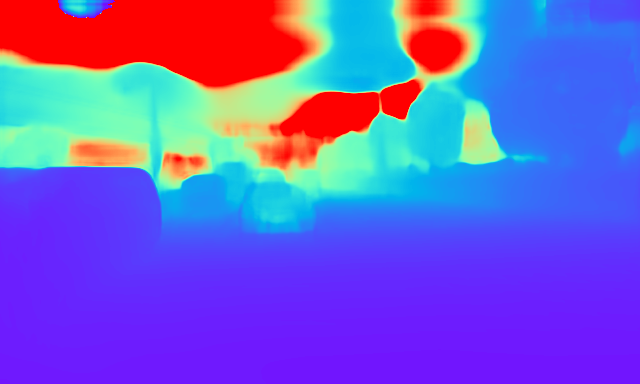}\hspace{-0.25em}
\includegraphics[width=2.9cm,height=1.74cm]{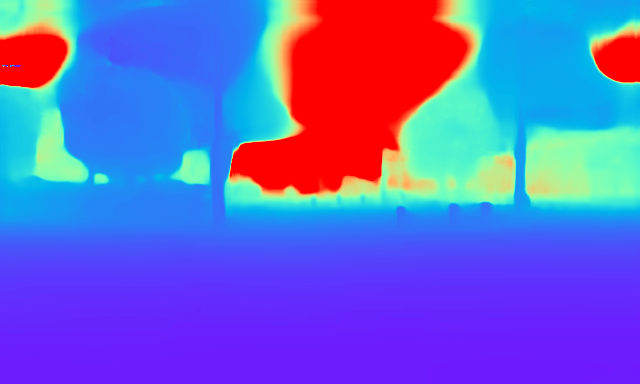}\hspace{-0.25em}
\includegraphics[width=2.9cm,height=1.74cm]{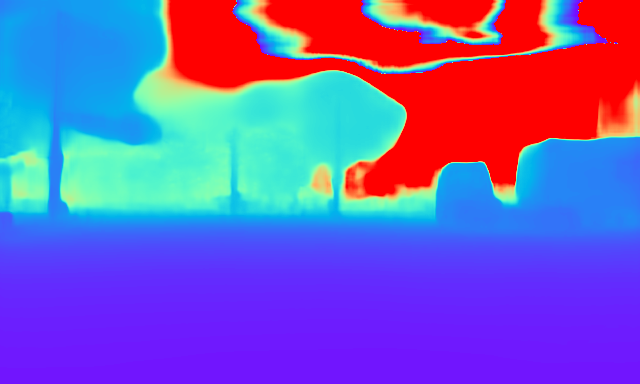}\hspace{-0.25em}
\includegraphics[width=2.9cm,height=1.74cm]{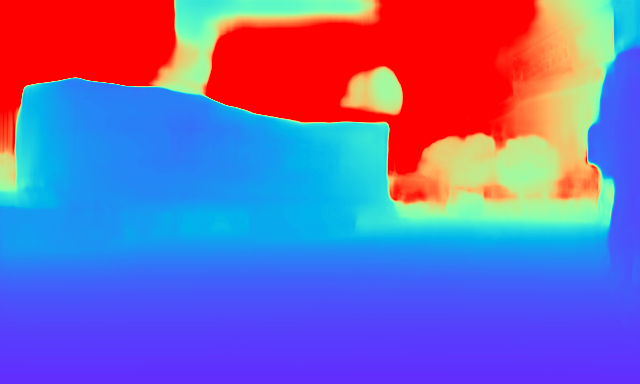}
}
\vspace{-0.8em}
\\
\subfloat{
\parbox[t]{2mm}{\rotatebox{90}{\small ~~FSM\cite{guizilini2022full}}}
\includegraphics[width=2.9cm,height=1.74cm]{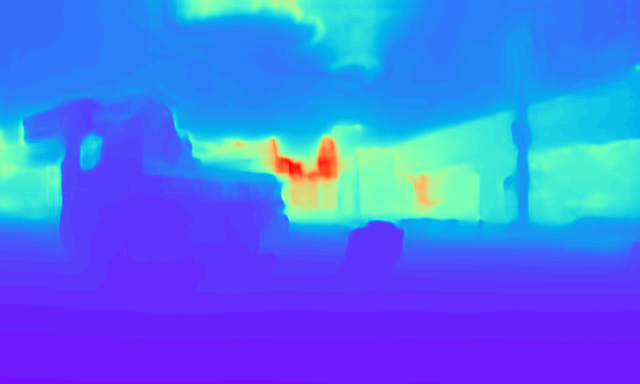}\hspace{-0.25em}
\includegraphics[width=2.9cm,height=1.74cm]{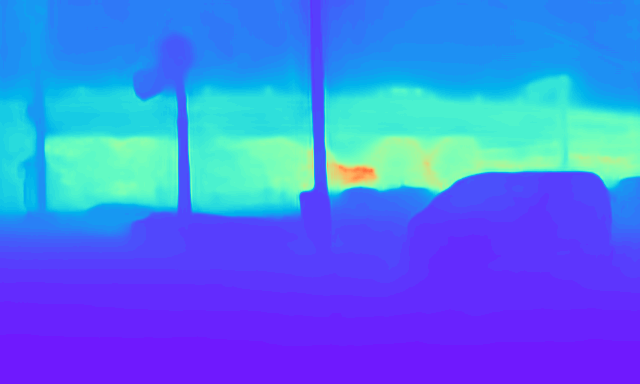}\hspace{-0.25em}
\includegraphics[width=2.9cm,height=1.74cm]{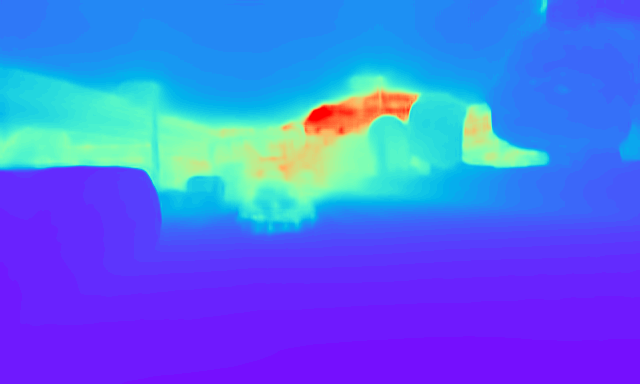}\hspace{-0.25em}
\includegraphics[width=2.9cm,height=1.74cm]{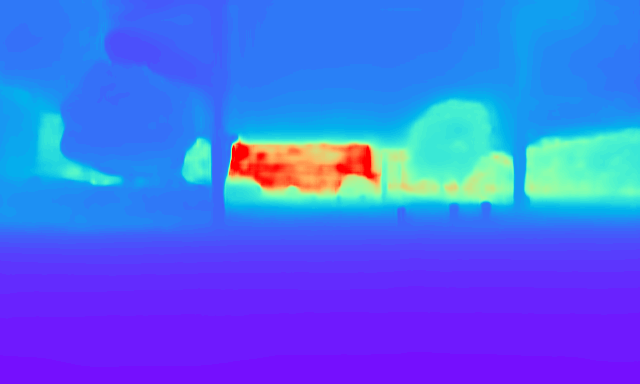}\hspace{-0.25em}
\includegraphics[width=2.9cm,height=1.74cm]{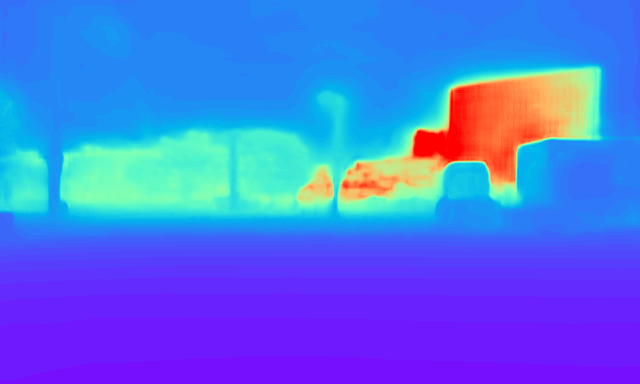}\hspace{-0.25em}
\includegraphics[width=2.9cm,height=1.74cm]{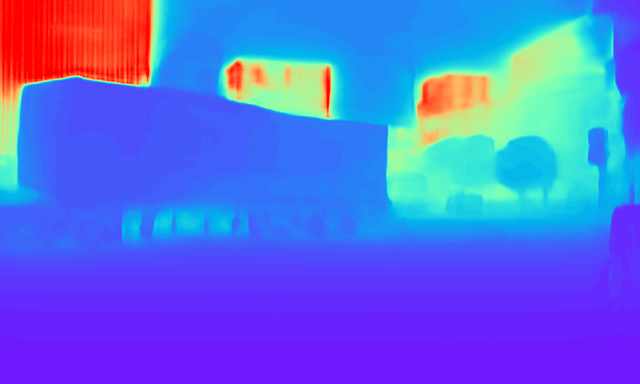}
}
\vspace{-0.8em}
\\
\subfloat{
\parbox[t]{2mm}{\rotatebox{90}{\small ~~~SD\cite{wei2022surrounddepth}}}
\includegraphics[width=2.9cm,height=1.74cm]{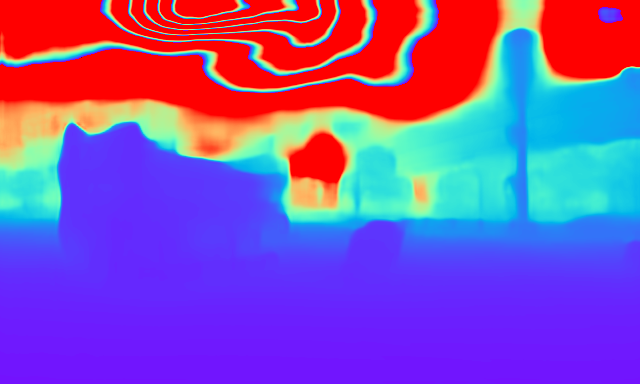}\hspace{-0.25em}
\includegraphics[width=2.9cm,height=1.74cm]{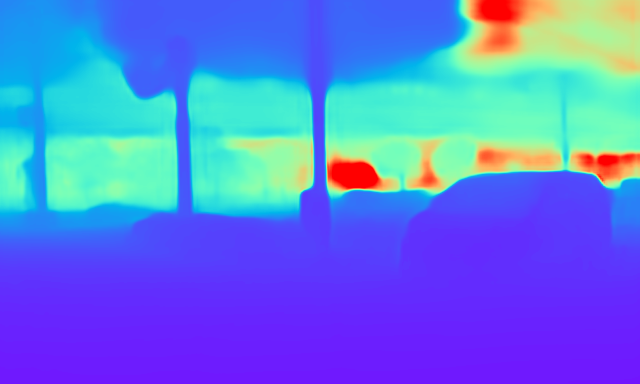}\hspace{-0.25em}
\includegraphics[width=2.9cm,height=1.74cm]{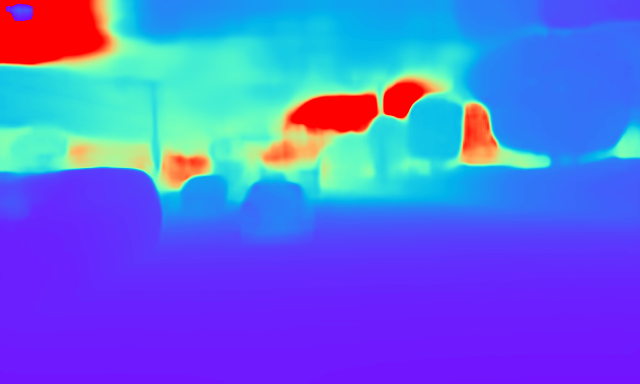}\hspace{-0.25em}
\includegraphics[width=2.9cm,height=1.74cm]{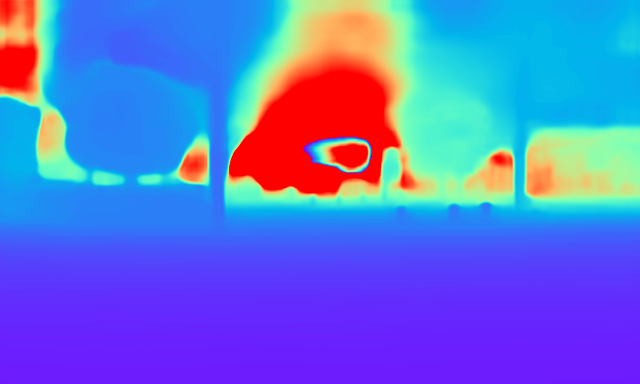}\hspace{-0.25em}
\includegraphics[width=2.9cm,height=1.74cm]{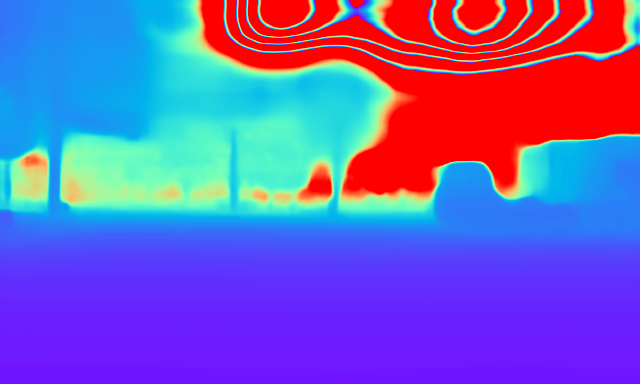}\hspace{-0.25em}
\includegraphics[width=2.9cm,height=1.74cm]{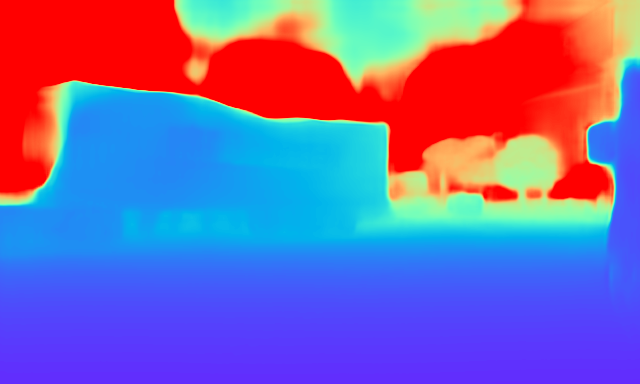}
}
\vspace{-0.8em}
\\
\subfloat{
\parbox[t]{2mm}{\rotatebox{90}{\small MCDP~\cite{xu2022multi}}}
\includegraphics[width=2.9cm,height=1.74cm]{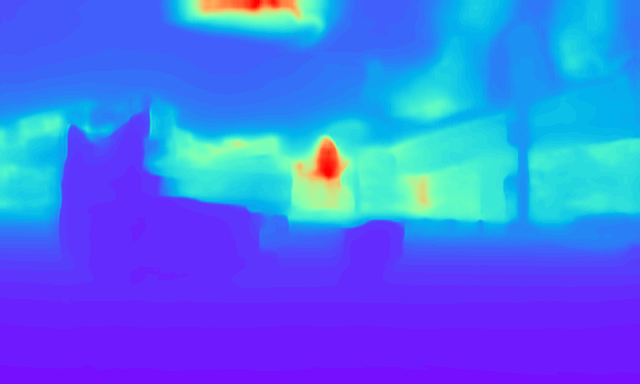}\hspace{-0.25em}
\includegraphics[width=2.9cm,height=1.74cm]{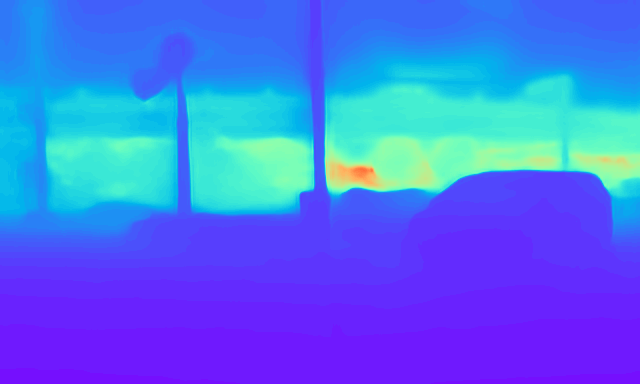}\hspace{-0.25em}
\includegraphics[width=2.9cm,height=1.74cm]{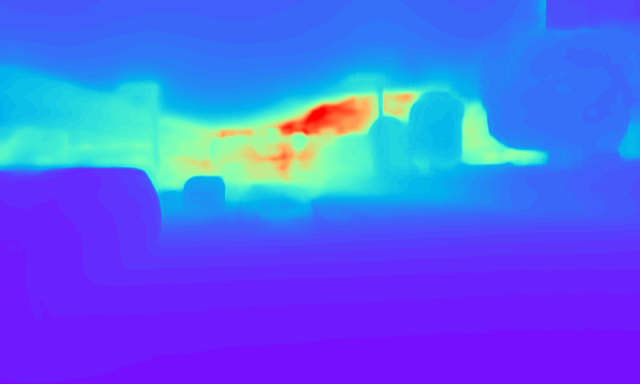}\hspace{-0.25em}
\includegraphics[width=2.9cm,height=1.74cm]{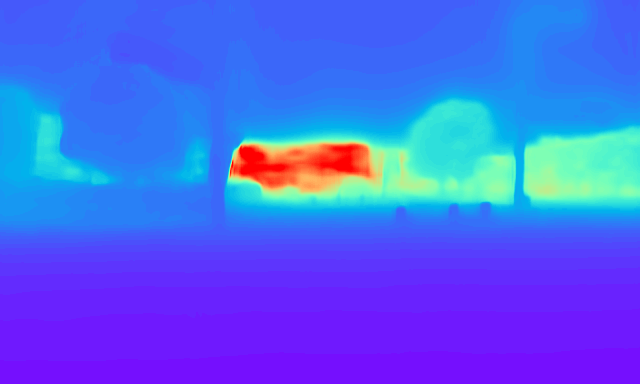}\hspace{-0.25em}
\includegraphics[width=2.9cm,height=1.74cm]{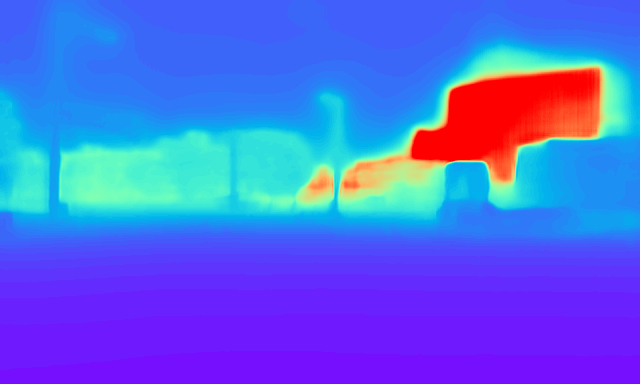}\hspace{-0.25em}
\includegraphics[width=2.9cm,height=1.74cm]{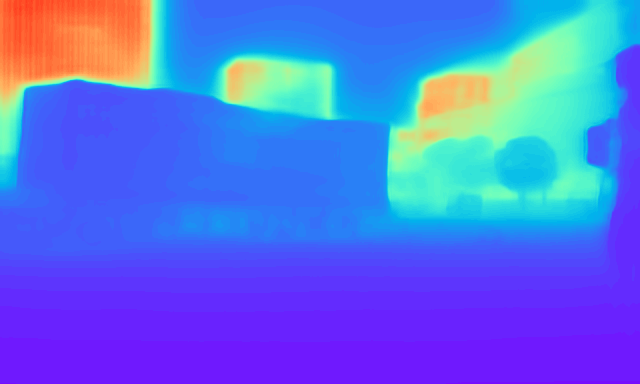}
}
\vspace{-0.8em}
\\
\subfloat{
\parbox[t]{2mm}{\rotatebox{90}{\small R18(\textit{i}$\&$L1)}}
\includegraphics[width=2.9cm,height=1.74cm]{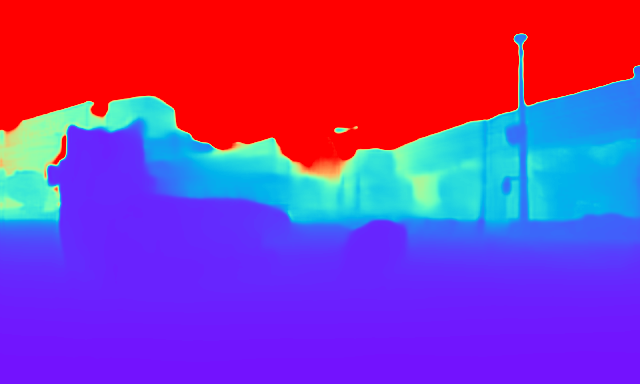}\hspace{-0.25em}
\includegraphics[width=2.9cm,height=1.74cm]{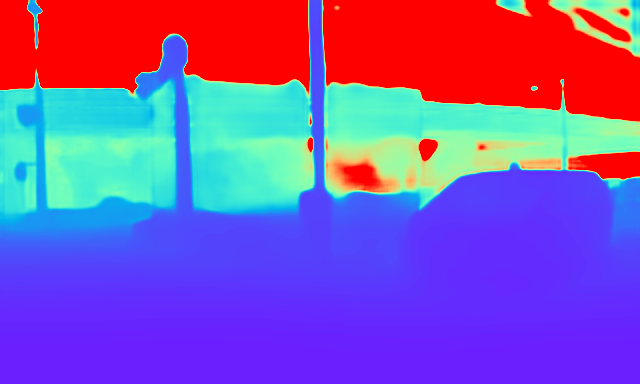}\hspace{-0.25em}
\includegraphics[width=2.9cm,height=1.74cm]{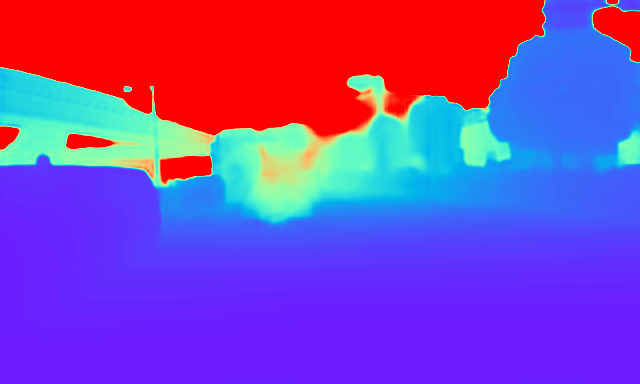}\hspace{-0.25em}
\includegraphics[width=2.9cm,height=1.74cm]{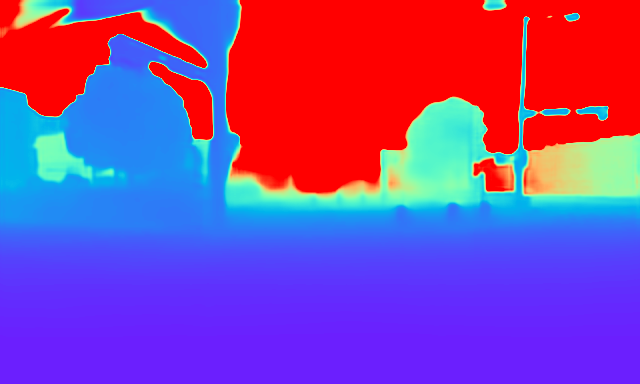}\hspace{-0.25em}
\includegraphics[width=2.9cm,height=1.74cm]{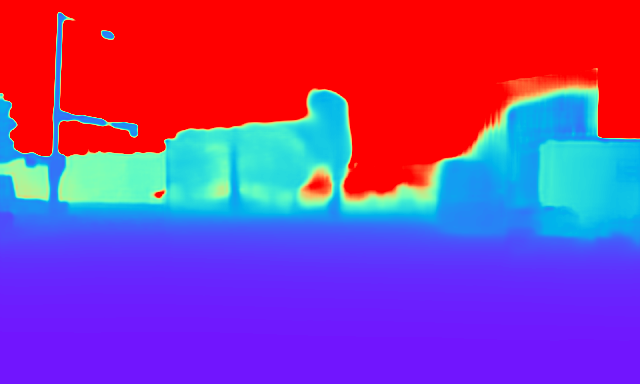}\hspace{-0.25em}
\includegraphics[width=2.9cm,height=1.74cm]{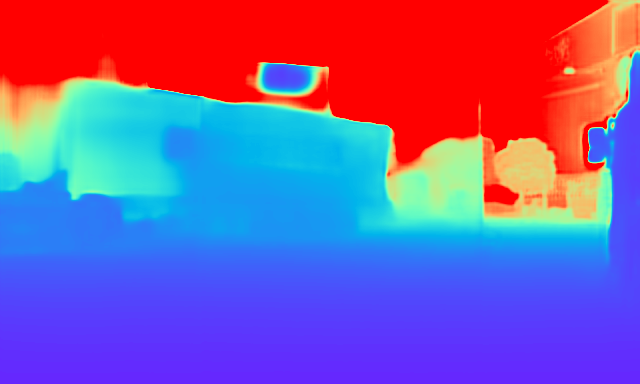}
}

\caption{
\textbf{Visualization of self-supervised depth prediction results on 6 cameras at the same frame in DDAD \cite{guizilini20203d}.}
Top (RGB): RGB pictures of six cameras. The red box denotes the overlapping region of each camera. The second row: the depth prior of each camera. The third row: results of the baseline R18~\cite{godard2019digging}. Fourth row: results of SurroundDepth~\cite{wei2022surrounddepth}. The last row (R18~(\textit{i} $\&$ {L1})): results of baseline R18~\cite{godard2019digging} applied with depth prior as input and L1 supervision.}

\label{fig:DDAD_qualitative}
\end{figure*}

 \label{sec:criteria}

\subsection{Depth Prediction with Depth Prior}
\label{sec:Depth Prediction}

With the virtual camera set up to solve the image quality problem of the fisheye camera, and obtaining accurate camera poses in real data,
we can now produce reliable depth prior $\mathbf{D}_p$ as described in Sec.~\ref{sec:stereo_rectify}. Then we use it to enhance the depth prediction.

First of all, $\mathbf{D}_p$ can be used as a part of the input to the model. Denote $f(\cdot|\boldsymbol{\theta})$ as the depth estimation model parametrized by $\boldsymbol{\theta}$, we predict the depth map $\mathbf{D}$:
\begin{equation}
\mathbf{D} = f(\mathbf{I}, \mathbf{D}_p | \boldsymbol{\theta}), 
\label{eq:input to model}
\end{equation}
where $\mathbf{I}$ is the color image. Due to its simplicity, this method can be applied to both supervised or self-supervised depth estimation models, and provide consistent improvements, as shown later in Sec.~\ref{sec:exp}. 

Due to the defects of the supervision signal, \textit{e.g.}, the sparsity of Lidar and the dynamic object problem in  temporal self-supervised signal, the loss on the depth prior can further provide extra supervision signals.
Specifically, in real scenes, we add extra loss terms into the training objectives during model training via
\begin{equation}
\underset{\boldsymbol{\theta}}{\text{minimize}}\ \  L_{ori}(\mathbf{D}, \mathbf{S}_{ori}) + \lambda L_{dp}(\mathbf{D}, \mathbf{D}_p)
\label{eq:l1 loss}
\end{equation}
where $L_{ori}(\cdot)$ and $\mathbf{S}_{ori}$ are the original loss and supervision signal of the model, and $L_{dp}(\cdot)$ is the extra supervision provided by the depth prior. We set the coefficient $\lambda$ to 0.005 in all experiments.
This method is inspired by~\cite{hu2021penet}, where the ground-truth depth is both given as inputs and supervision signals to form a depth completion task and boost the depth prediction accuracy. 
We experiment in Sec.~\ref{sec:exp} with different realizations of $L_{dp}(\cdot)$, including the L1 loss, the virtual normal loss~\cite{Yin_2019_ICCV} and the ranking loss ~\cite{Xian_2020_CVPR}, which all provide effective improvements and verify our hypothesis. 

\begin{table*}[t]
\renewcommand\arraystretch{1.5}
\setlength\tabcolsep{20pt} 
\begin{tabular}{c|c|ccccc}
    \toprule
Supervision
    &Method& $>$1 $\downarrow$ & $>$3 $\downarrow$ & $>$5 $\downarrow$ & MAE $\downarrow$ & RMS $\downarrow$  \\ 
%
\midrule
\multirow{1}{*}{Unsupervised} & RAFT-Stereo& 19.29 
& 11.485 
& 9.572
& 1.223
& 3.482
\\
\midrule
\multirow{5}{*}{Ground-truth}&PSMNet~\cite{chang2018pyramid} &92.67 & 31.45 & 21.32 & 4.33 & 7.76\\

&DispNet~\cite{Ilg_2018_ECCV}
   &39.02 & 21.12 & 14.47 & 2.37 & 4.85 \\
\cline{2-7}

&Omni~\cite{won2019omnimvs}
    &14.102
    & 4.014
    & 2.644 
    & 0.826
    & 2.453
    \\
 & +RAFT input
    &13.482
    & 3.797
    & 2.524
    & 0.788
    & 2.355
    \\
 & +SGBM input
    &\textbf{12.223}
    & \textbf{3.620}
    & \textbf{2.352}
    & \textbf{0.735}
    & \textbf{2.298}
    \\
\bottomrule
\end{tabular}
\caption{\label{tab:fisheye}{\textbf{Performance of SGDE in Synthetic Urban dataset~\cite{won2020end}.}  Unsupervised denotes the depth prior is inferred by RAFT-Stereo~\cite{teed2020raft}} without re-training. SGBM~\cite{hirschmuller2005accurate} and RAFT (RAFT-Stereo~\cite{lipson2021raft}) denotes the depth prior is computed by the corresponding methods.}

\end{table*}

\begin{table}[t]%

\centering
\small
\renewcommand{\arraystretch}{1.05}
\setlength{\tabcolsep}{0.22em}
\begin{tabular}{l|l|cccc}
\toprule
\multirow{1}{*}{Dataset} & \multirow{1}{*}{Pose}  &Abs Rel$\downarrow$ & Sq Rel$\downarrow$ &RMSE$\downarrow$ & $\delta_{1.25}$ $\uparrow$   \\

\midrule
\multirow{4}{*}{DDAD~\cite{guizilini20203d}}
& GT            & 0.885 & 8.392 & 24.184 &  0.439\\
& HOM~\cite{ling2016high}           & 0.599 & 6.381 & 18.209 &  0.498\\
&w/o GLC & 0.474 & 5.827 & 15.230 &  0.565\\
&GLC &  \textbf{0.142} &\textbf{2.331} &\textbf{10.103}&\textbf{0.803}  \\ 

\midrule
\multirow{4}{*}{nuScenes~\cite{caesar2020nuscenes}}
& GT            & 0.940 & 6.129 & 13.113 & 0.410  \\
& HOM~\cite{ling2016high}           & 0.885 & 5.827 & 12.609 & 0.446  \\
&w/o GLC & 0.779 & 5.523 & 10.993 & 0.511  \\
&GLC & \textbf{0.498} &\textbf{3.931} & \textbf{9.102} & \textbf{0.597}  \\ 
\bottomrule
\end{tabular}
\caption{\textbf{Effect of pose optimization on the accuracy of the depth prior}. The ``ground truth" camera poses provided in the datasets (GT) are not accurate enough for geometry methods, resulting in high depth reconstruction error. The SOTA stereo calibration method HOM~\cite{ling2016high} can not reduce the pose error effectively. 
Using geometric loop constraint (GLC) to calibrate cameras can effectively reduce the error of depth prior than not using it (w/o GLC). 
}
\label{tab:overlap result}
\end{table}

\begin{table}[t]
\small
\renewcommand{\arraystretch}{1.05}
\setlength{\tabcolsep}{0.30em}

\subfloat[\small{DDAD}]{
\begin{tabular}{l|cccc}
    \toprule
    Method
    
    &Abs Rel$\downarrow$ & Sq Rel$\downarrow$ &RMSE$\downarrow$ & $\delta_{1.25}$ $\uparrow$   \\ 
\midrule
Baseline (R18)~\cite{godard2019digging}~~~~~~~
    & 0.167 & 2.718 & 11.302 & 0.773  \\
+ input
    & 0.146 & 2.342 &  10.468 & 0.807 \\
+ L1 
    & 0.162 & 2.705 & 11.300 & 0.774\\
+ input $\&$ L1
    & \textbf{0.144} & \textbf{2.295}  & \textbf{10.165} & \textbf{0.813}\\
\midrule
Adabins~\cite{bhat2021adabins}
    & 0.156 & 2.511 & 10.864 & 0.808  \\
+ input
    & {0.136} & {2.203} & {10.001} & {0.815}  \\
+ input \& L1
& \textbf{0.129} & \textbf{2.178} & \textbf{9.864} & \textbf{0.824}  \\
    \bottomrule
\end{tabular}
}
\\

\subfloat[\small{nuScenes}]{
\begin{tabular}{l|cccc}
  \toprule
    Method
    
    &Abs Rel$\downarrow$ & Sq Rel$\downarrow$ &RMSE$\downarrow$ & $\delta_{1.25}$ $\uparrow$   \\ 
\midrule

Baseline (R18)\cite{godard2019digging}~~~~~~~

    &0.113 & 1.266 & 3.332 & 0.914  \\

+ input
    &{0.095} & {1.124} & {2.937} & {0.936}  \\
+ input  \& L1
    &\textbf{0.089} & \textbf{1.118} & \textbf{2.925} & \textbf{0.939}  \\
\midrule
Adabins~\cite{bhat2021adabins}

    &0.125 & 0.910 & 4.115 & 0.866\\
+ input
        &{0.112} & {0.861} & {3.700} & {0.882}    \\
+ input \& L1
        &\textbf{0.104} & \textbf{0.847} & \textbf{3.568} & \textbf{0.892}    \\
\bottomrule
\end{tabular}
}

\caption{\label{tab:supervised-result}\textbf{Performance of SGDE for supervised depth estimation.} L1 denotes the L1 loss on depth prior. Depth prior is computed by RAFT-Stereo~\cite{lipson2021raft}.
}
\end{table}


\section{Experiments}\label{sec:exp}

\subsection{Datasets}

\noindent\textbf{Synthetic Urban~\cite{won2019sweepnet}}.
Synthetic Urban is virtually implemented with the camera rig by four fisheye cameras, and the dataset is rendered using Blender. The FOV of each camera is 220\degree. The dataset contains three scenes: sunny, cloudy and sunset, and each consists of 1000 sequential frames of city landscapes. We use the sunny part in our experiments. 

\noindent\textbf{DDAD \cite{guizilini20203d}}. 
Dense Depth for Automated Driving (DDAD) is an urban driving dataset captured by six calibrated cameras time-synchronized at 10 Hz with relatively small overlaps. There are 200 sequences with different calibration parameters including 12650 training samples (63250 images) and 3950 validation samples (15800 images). 

\noindent\textbf{nuScenes \cite{caesar2020nuscenes}}. The nuScenes is an urban driving dataset containing images from a synchronized six-camera array at 12Hz. The calibration  provided by dataset is only done on keyframes at 2 Hz and 
we optimize the poses at 12 Hz. It consists of 1000 scenes with 1.4 million images.
%

\subsection{Implementation details}

We rely on OpenCV~\cite{bradski2000opencv} and PyTorch~\cite{paszke2019pytorch} for implementations. 
Following~\cite{guizilini2022full}, the original image is downsampled before given to the network. The resulting resolution is 640 $~\times$ 384 in DDAD and 768 $\times$ 448 in nuScenes. For evaluation, we use bilinear interpolation to upsample the predicted depth images to the original resolution. 
We annotate occlusion masks manually following ~\cite{guizilini2022full}, which are used to remove wrong depth estimation results caused by the car body occlusion, and to reduce errors in the training model. 
We use RAFT-Stereo~\cite{lipson2021raft} as the default stereo matching method.
We use the original hyper-parameters when applying SGDE to existing models. 
In DDAD and nuScenes datasets, we optimize the camera pose once per sequence by the first 7 frames and the optimized pose is applied to the whole sequence.
For a fair comparison, in Tab.~\ref{tab:self-supervsied DDAD}, \ref{tab:unsupervised-result nuscenes}, \ref{tab:supervised-result} and \ref{tab:Depth consistency} the experimental results are calculated by the optimized pose.
Due to the problems (i.e. uneven pixel density and distortion) at the edge of the fisheye image are very serious, we discard a few pixels at the edge of the images during transforming the fisheye camera to the pinhole camera. Therefore, the depth priors in fisheye cameras are a little bit smaller than RGB, and most of the valuable areas are preserved as shown in Fig. \ref{fig:fisheye}.

In surrounding view calibration, we establish the geometric loop constraint (GLC) given the video sequence with $T$ = 7 frames (10 Hz in DDAD dataset~\cite{guizilini20203d}, 12 Hz in nuScenes dataset~\cite{caesar2020nuscenes}). The nuScenes dataset only provides the calibration of keyframes at 2 Hz, and we choose two keyframes and five non-keyframes between them to optimize the pose. 
The initial poses of keyframes are set to the ground-truth provided by dataset, and we initial the pose calibration of non-keyframes before the bundle adjustment stage.  
Due to the small overlap area, the scene complexity and the small value of $T$ (set to 7 in all experiments for efficiency) we cannot guarantee the success of optimization on every small sequence. Therefore, we introduce effective criteria to filter out bad optimization results. Firstly, we discard all correspondences between two observations if their number is fewer than 
$\beta$ ($\beta$ = 200 in DDAD~\cite{guizilini20203d}, $\beta$ = 50 in nuScenes~\cite{caesar2020nuscenes}). 
Secondly, during bundle adjustment, when the maximum number of iterations 200 is reached yet the optimization has not converged, the optimization is abandoned. 
Finally, if the translation distance of the relative pose 
before and after optimization exceeds $\alpha = 0.3 m$, the optimization result is also discarded. In all abandoned or not-optimized video sequences, relative camera poses maintain the last successful optimization result based on the timestamps.
The operation of calculating depth prior is on the rectified camera plane and we need to project it back to the original camera plane.
As input, the depth prior concatenates with RGB in channels. The initial weight of the depth prior channel is the average of the initial weight value of the RGB three channels. We normalize the  depth prior to 1 and fill the non-overlap areas with 0.


\begin{table*}[t]
\centering
\small
\renewcommand\arraystretch{1.5}
\setlength\tabcolsep{20pt} 

\begin{tabular}{l|c|cccc}
\toprule
\multicolumn{6}{c}{Abs Rel $\downarrow$}\\
\midrule
Method & Stereo Method & W/O $\mathbf{D}_p$ & + input & + L1 & + input $\&$ L1    \\ 
\midrule
SC-Depth~\cite{bian2019unsupervised} & -&\multicolumn{1}{c}{0.210} & -&  - &- \\
SC-Depth + Ours & RAFT &- & 0.185  & 0.201  & \textbf{0.178}\\
\midrule

PackNet~\cite{guizilini20203d} &- &\multicolumn{1}{c}{0.215} & -  & - &-  \\
PackNet + Ours & RAFT &-&  0.188 & 0.206 & \textbf{0.177} \\
\midrule

FSM~\cite{guizilini2022full} &-& 0.202   -  &- &-  & -\\
FSM + Ours & RAFT &   - & 0.185 &  0.191 &  \textbf{0.178}\\
\midrule
MCDP~\cite{xu2022multi} &-& 0.193 &  -& -  &- \\
MCDP + Ours & RAFT &  -& 0.180 & 0.187  & \textbf{0.177}\\
\midrule
SurroundDepth~\cite{wei2022surrounddepth} &-& 0.200
     & - & - &- \\
SurroundDepth + Ours & RAFT & -
& 0.181 & 0.190 & \textbf{0.175}\\
\midrule
BEVScope~\cite{mao2023bevscope}  & - &  0.191 & - & - & - \\
BEVScope + Ours & RAFT & - & 0.179 & 0.185 & \textbf{0.174} \\
\midrule
Baseline (R18)   & -& \multicolumn{1}{c}{0.220}
     & -  & - & -  \\
Baseline (R18) + Ours  & SGBM & 
     - & 0.191  & 0.213  & 0.184\\
Baseline (R18) + Ours & RAFT & - & 0.186  & 0.209  & \textbf{0.178}  \\

\bottomrule
\end{tabular}

\caption{\label{tab:self-supervsied DDAD}{\textbf{Abs Rel of self-supervised depth results in DDAD.} 'w/o $\mathbf{D}_p$' denotes the results of each method without depth prior. input denotes using depth prior as input and L1 denotes the L1 loss on depth prior. SGBM~\cite{hirschmuller2005accurate} and RAFT (RAFT-Stereo~\cite{lipson2021raft}) denotes the depth prior is computed by the corresponding methods.}}

\end{table*}
\subsection{Benchmark and Evaluation Metrics}

We evaluate the generality of SGDE by applying it to both pinhole and fisheye camera models in both self-supervised and supervised depth estimation tasks.

In Synthetic Urban dataset, we follow ~\cite{won2019omnimvs} to evaluate the performance of SGDE in existing supervised SOTA models~\cite{won2019omnimvs,chang2018pyramid,Ilg_2018_ECCV}. And there is no method achieving unsupervised depth estimation without temporal information. We evaluate the performance by mean absolute error (MAE), root mean squared error (RMSE) and \textquotesingle $>$n \textquotesingle, where {\textquotesingle $>$n \textquotesingle} refers to the pixel ratio whose error is large than n.

In DDAD and nuScenes datasets, following~\cite{guizilini2022full} in the self-supervised task, we choose Mono2~\cite{godard2019digging} with Resnet-18 (R18) as the baseline, and include two monocular methods PackNet~\cite{guizilini20203d} and SC-Depth ~\cite{bian2019unsupervised}. We also evaluate three multi-camera depth estimation methods FSM~\cite{guizilini2022full}, MCDP~\cite{xu2022multi}, 
and SurroundDepth~\cite{wei2022surrounddepth}. For supervised depth prediction, there is no prior work focusing on our setup. Using the depth ground-truth provided by the dataset as supervision, we evaluate the baseline Resnet-18 (R18) model (the depth network in ~\cite{godard2019digging})  and the SOTA model Adabins~\cite{bhat2021adabins}.
We use the standard metrics to evaluate the quality of the depth prediction: average relative error (Abs Rel), squared relative difference (Sq Rel), root mean squared error (RMSE), threshold accuracy ($\delta_{1.25}$). In addition, we evaluate the consistency of depth estimates in overlap regions by the depth consistency metric (Dep Con)~\cite{xu2022multi}.  Dep Con project the predicted depth from the overlapping regions of adjacent cameras onto the same plane and calculate the error with the ground-truth.
We train one model to estimate the depth for all cameras. The distances are up to 200m in DDAD and 80m in nuScenes.


\begin{table*}[t]
\centering
\small
\renewcommand\arraystretch{1.5}
\setlength\tabcolsep{18pt} 

\begin{tabular}{l|cccccc}
  \toprule
    Method
    &Abs Rel$\downarrow$ & Sq Rel$\downarrow$ &RMSE$\downarrow$ & $\delta_{1.25}$ $\uparrow$ & $\delta_{1.25}^2$ $\uparrow$ &$\delta_{1.25}^3$ $\uparrow$   \\ 
\midrule
PackNet~\cite{guizilini20203d} &0.303 & 3.154 & 7.014 & 0.655 & 0.822 & 0.901  \\
~~~~+ stereo input     & \textbf{0.279} & \textbf{3.001}& \textbf{6.920}&\textbf{0.711}&\textbf{0.853}&\textbf{0.926} \\
\midrule
FSM~\cite{guizilini2022full}  &0.271 & 3.187 & 6.832 & 6.688 & 0.840 & 0.895\\
~~~~+ stereo input    & \textbf{0.251} & \textbf{3.023} & \textbf{6.810}  & \textbf{7.121} & \textbf{0.852} & \textbf{0.932} \\
\midrule
MCDP~\cite{xu2022multi}    &0.238 & 3.030 & 6.823 & 0.717  & 0.880 & 0.924 \\
~~~~+ stereo input    & \textbf{0.229} & \textbf{3.012} & \textbf{6.803} & \textbf{0.721}  & \textbf{0.891}  & \textbf{0.946}\\
\midrule 
SurroundDepth~\cite{wei2022surrounddepth}    &0.246 & 3.069 & 6.835 & 0.719 & 0.879 &  0.936 \\
~~~~+ stereo input    &\textbf{0.225} & \textbf{2.987} & \textbf{6.756} & \textbf{0.728} &  \textbf{0.887} & \textbf{0.945}  \\
\midrule
BEVScope~\cite{mao2023bevscope} & 0.232 & 2.652 & 6.672 & 0.720 & 0.876 & 0.936 \\
~~~~+ stereo input & \textbf{0.232} & \textbf{2.652} & \textbf{6.672} & \textbf{0.720} & \textbf{0.888} & \textbf{0.954} \\
\midrule

Baseline (R18)~\cite{godard2019digging}
    &0.308 & 3.460 & 8.014 & 0.659  & 0.851 & 0.927 \\
~~~~+ stereo input
    &0.289 & 3.212 & 7.249 & 0.691  & 0.872 & 0.952 \\
~~~~+ stereo input $\&$ VN
    & 0.270 & 3.102 & 7.058 & 0.704  & 0.888 & 0.959\\
~~~~+ stereo input $\&$ RL
    & 0.276 & \textbf{3.029} & 7.010 & 0.700 & 0.886 &  0.964  \\
~~~~+ stereo input $\&$ L1
    &\textbf{0.268} & 3.190  & \textbf{6.949} & \textbf{0.705}  & \textbf{0.891} & \textbf{0.966}\\
\bottomrule

\end{tabular}

\caption{\label{tab:unsupervised-result nuscenes}{\textbf{Performance of self-supervised depth estimation results in nuScenes.}  VN, RL, L1 denote the Virtual Normal loss~\cite{Yin_2019_ICCV}, Ranking loss~\cite{Xian_2020_CVPR} and L1 loss on depth prior.  Depth prior is computed by RAFT-Stereo~\cite{lipson2021raft}.
} }

\end{table*}


\subsection{Main Result}


As shown in Tab. \ref{tab:fisheye}, \ref{tab:overlap result},  the accuracy of unsupervised depth prior surpasses that of some Lidar supervised methods in Synthetic Urban and DDAD datasets. We also show the visualization of the depth prior in Fig. \ref{fig:fisheye} and \ref{fig:DDAD_qualitative}.

As shown in Tab.~\ref{tab:fisheye}, \ref{tab:self-supervsied DDAD}, \ref{tab:unsupervised-result nuscenes},  and \ref{tab:supervised-result}, SGDE can be widely used in fisheye and pinhole camera models. And in \emph{both} self-supervised and supervised tasks, SGDE significantly and consistently improves the baselines and SOTA models. 
Applying the depth prior as an extra input (+ input) provides effective improvements. 
Adding depth prior as extra supervision signals (VN/RL/L1) can further improve the performance.
We also show in Fig.~\ref{fig:DDAD_qualitative} the qualitative results of different methods in DDAD dataset~\cite{guizilini20203d}.
With the help of SGDE, the Resnet-18 base model produces sharper and more accurate results than its baseline and SOTA method.

\begin{table}[t]%

\centering
\small
\renewcommand{\arraystretch}{1.05}
\setlength{\tabcolsep}{0.22em} 
\begin{tabular}{l|cccccc}
\toprule
\multirow{2}{*}{Method} &
\multicolumn{6}{c}{Dep Con$\downarrow$} \\

\cmidrule{2-7}
 & \textit{Front} & \textit{F.Left} & \textit{F.Right} & \textit{B.Left} & \textit{B.Right} & \textit{Back} \\

\midrule
FSM~\cite{guizilini2022full}  & 0.731  & 0.382  & 0.331   &  0.460  &  0.381 &  0.511 \\
SD~\cite{wei2022surrounddepth} &  {0.740}   & {0.392}    & {0.320}    & {0.455}     &{0.390}     & {0.698}  \\
MCDP~\cite{xu2022multi} & 0.729 & 0.386 & 0.315 & 0.457 & 0.371 & 0.537 \\

\midrule
Baseline (R18)~\cite{godard2019digging}~~~~
& 0.752  & 0.451  & 0.348   &  0.468  &  0.388 &  0.702  \\
+ input
& 0.702  & 0.382  & 0.309   & 0.436  &  0.362 &  0.628 \\
+ input $\&$ L1
& \textbf{0.697}  &\textbf{0.377}  & \textbf{ 0.293}   & \textbf{0.428}   & \textbf{ 0.311} & \textbf{0.451}   \\
\bottomrule
\end{tabular}

\caption{\textbf{Depth consistency results of each camera in DDAD}. \emph{Dep Con}~\cite{xu2022multi} is the standard metric to evaluate depth consistency in overlap areas. Applying SGDE on the baseline achieves the SOTA results.
}
\label{tab:Depth consistency}

\end{table}

\begin{figure}[t]
\centering

\includegraphics[width=8.0cm]{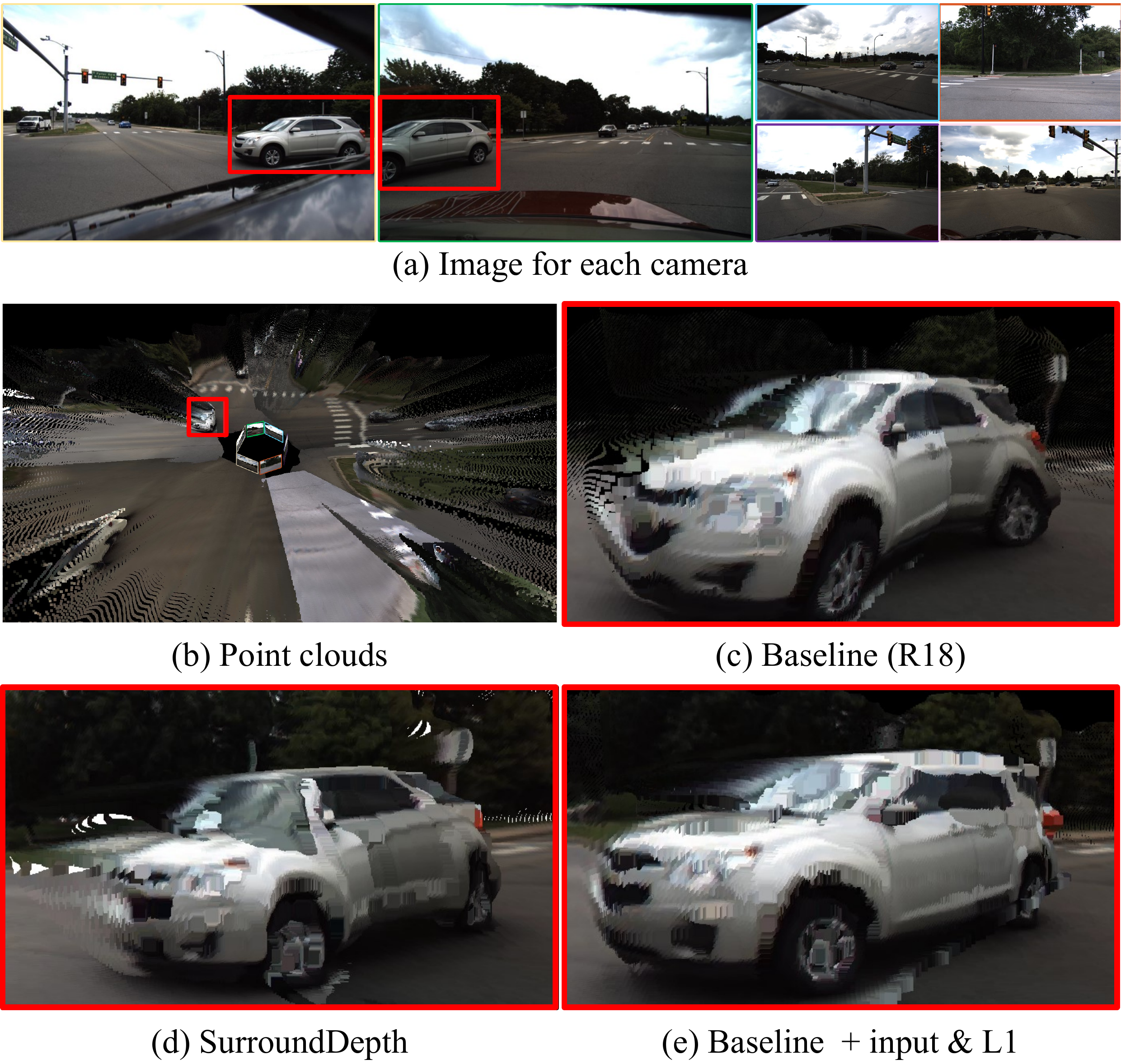}
%
\caption{\textbf{Self-supervised predicted point cloud result comparison on DDAD dataset.} (a) is multiple camera views in the same frame. (b) is the predicted surrounding point clouds. The car appearing in the overlapping area is enlarged in (c), (d) and (e) by different methods.}
\label{fig:pc}
\end{figure}

\subsection{Analysis}

\subsubsection{Pose Optimization Matters}
We evaluate the accuracy of the depth prior calculated from the different calibration methods.
As shown in Tab.~\ref{tab:overlap result}, using inaccurate ``ground-truth" poses (GT) resulted in poor depth prior. Because of the small overlap area between cameras, the existing SOTA stereo camera calibration method HOM~\cite{ling2016high} can not effectively reduce the error. 
With geometric loop constraint (GLC), our surrounding view calibration method can effectively reduce the error of the depth prior.
In order to ablate the role of the proposed geometry loop constraint in our calibration method, we perform our calibration scheme on two adjacent cameras, which cannot build the geometry loop constraint.
As shown in Tab.~\ref{tab:overlap result}, experimental results show that compared with our method using geometry loop constrain (GLC), the error of depth prior is significantly increased due to the lack of that (w/o GLC).
This result is consistent in both datasets and proves that the geometry loop constraint is indeed  effective in the surrounding view calibration method.

\subsubsection{Depth Consistency in Overlap Regions}
The consistency of the predicted depth on the overlap regions is an important metric for downstream tasks such as 3D reconstruction. We show in Fig.~\ref{fig:pc} the point cloud predicted by the Resnet-18 based self-supervised method on six cameras in DDAD and purposely zoom in on a car in the overlapping area. In Fig.~\ref{fig:pc}{c} and ~\ref{fig:pc}{d}, the car is heavily distorted, which is caused by the cross-view depth prediction inconsistency. After applying SGDE, the car becomes smooth and consistent in Fig.~\ref{fig:pc}{e}. We also follow MCDP~\cite{xu2022multi} to evaluate the depth consistency by \emph{Dep Con} as shown in Tab.~\ref{tab:Depth consistency}.
Compared with the baseline model,
SGDE greatly improves the depth consistency across cameras, achieving new state-of-the-art performance.

\subsubsection{Stereo Matching Methods}
We evaluate the effect of the different stereo matching algorithms  in SGDE.
We compare the hand-crafted method SGBM~\cite{hirschmuller2005accurate} with learning based method RAFT-Stereo~\cite{lipson2021raft}. As shown in Tab.~\ref{tab:fisheye} and \ref{tab:self-supervsied DDAD}, SGBM also works in our SGDE pipeline, and this non-parametric approach imposes little burden on practical applications, but the benefits are great.

\subsubsection{Challenges in Establishing Spatio-Temporal Matches}

We argue that the 360\degree camera ring set often has limited and low-quality overlap regions, making multi-view stereo (MVS) methods infeasible for the entire image. As an offline task, it is a common practice to establish spatio-temporal matches between different cameras at different times to increase the overlap area. For example, the back-right camera at time $T$ matches with the front-right camera at time $T-K$, for some value of $K$. 
However, this strategy faces significant limitations in autonomous driving applications.

Spatio-temporal matches assumes that the entire scene is static, but dynamic objects (\textit{i.e.}, pedestrians, and vehicles) have different positions at different frames, which causes the method of MVS to fail. And dynamic objects are important to observe in autonomous driving applications.  To test this idea, we use the image of one camera at the $T$ frame, and the image of the adjacent camera at the $T+10$ frame to perform stereo matching in DDAD dataset. We use 2D segmentation maps from off-the-shelf segmentors~\cite{zhang2023simple} to label the dynamic objects (pedestrians,  bicycles, and vehicles).  As shown in Tab.~\ref{tab:matching method}, 
spatio-temporal matching introduces significant errors in the depth estimation of dynamic objects (pedestrians, bicycles, and vehicles), but our method is not affected by them.

\begin{figure*}[t]
\flushleft
\centering
\subfloat{
\includegraphics[width=18cm]{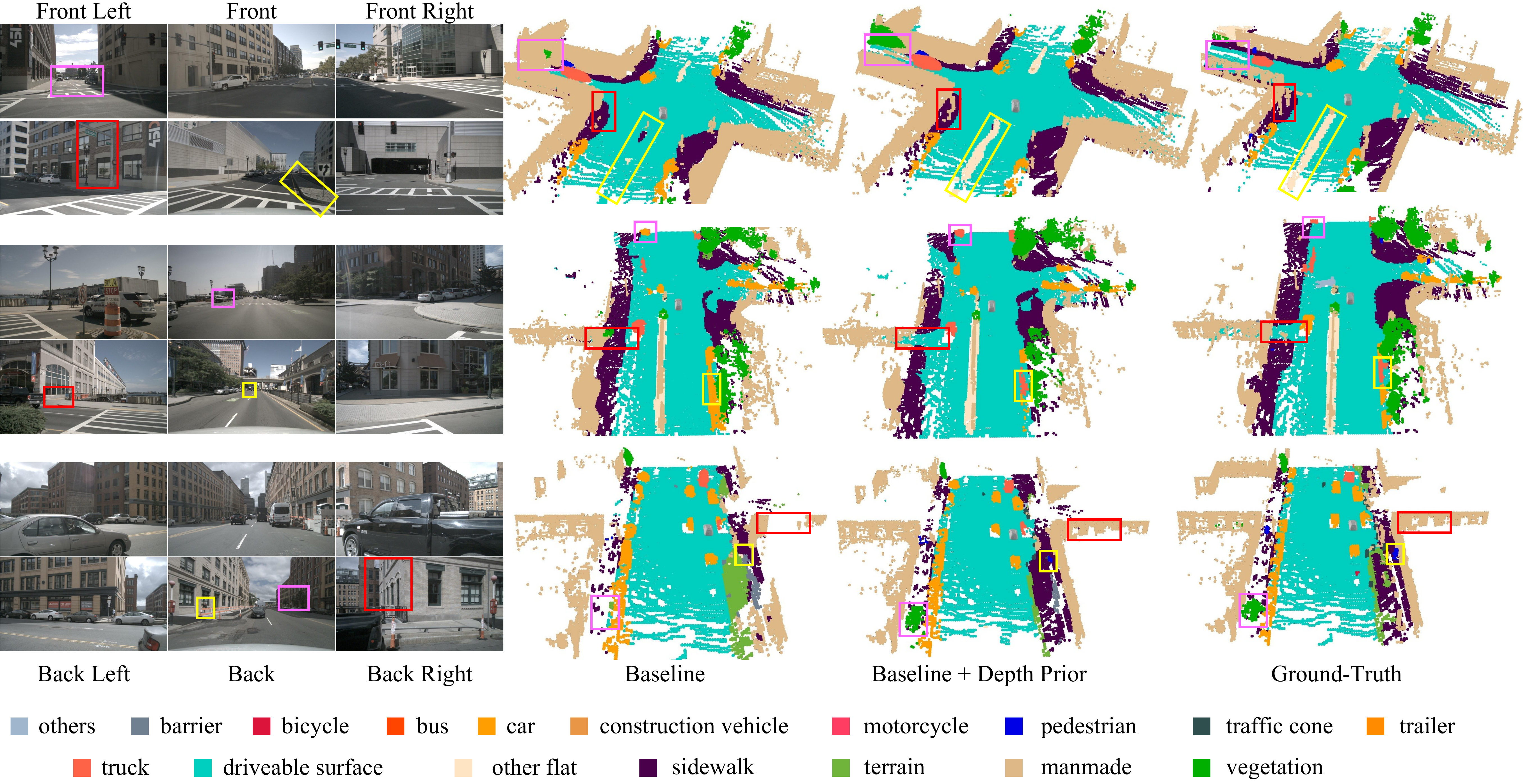}
} 
\caption{{
Visualization comparison for 3D semantic occupancy prediction  between baseline and adding depth prior.} We use VoxFormer~\cite{li2023voxformer} as baseline  and add depth prior as input and supervisior on depth estimation branch on the Occ3D-nuScenes~\cite{tian2023occ3d} dataset.}
\label{fig:occ}
\end{figure*}

\begin{table}[t!]
\centering
\renewcommand\arraystretch{1.5}
\setlength\tabcolsep{7pt}
\begin{tabular}{c|c|c}
\hline
Method & Static Scenes & Dynamic Objects \\ 
\hline
Spatio-temporal Match   & 0.145 & 0.438 \\
Match in the same frame (Ours)            & \textbf{0.144} & \textbf{0.136} \\
\hline
\end{tabular}
\caption{\label{tab:matching method}{
The Abs Rel Error of stereo-matching (RAFT) under different matching schemes. We emphasize performance on static scenes and dynamic objects (pedestrians,  bicycles, and vehicles).}}
\end{table}

\begin{table}[t!]
\centering

\renewcommand\arraystretch{1.5}
\setlength\tabcolsep{7pt}
\begin{tabular}{c|c|c|c}
\hline
Method & Abs Rel $\downarrow$ & FPS $\uparrow$ & GPU Memory $\downarrow$ \\ 
\hline
SurroundDepth   & 0.200 & 26.8 & 9.2 GB \\
MCDP            & 0.193 & 28.9 & 8.8 GB \\
BEVScope        & 0.191 & 17.8 & 11.3 GB\\
\hline
Baseline (R18)  & 0.220 & \textbf{37.8} & \textbf{5.4 GB} \\
Baseline + Ours(SGBM) & 0.184 & 36.6 & {5.7 GB} \\
Baseline + Ours(RAFT) & \textbf{0.178} & 33.4 & 8.5 GB\\
\hline
\end{tabular}

\caption{\label{tab:FPS}{The results of Abs Rel error, FPS, and GPU memory of self-supervised depth estimation in DDAD dataset. 
}}
\end{table}

\begin{table*}[t]
\centering
\small
\renewcommand\arraystretch{1.5}
\setlength\tabcolsep{12pt} 

\begin{tabular}{l|ccccccc}
  \toprule
    Depth estimation Method
    &mAP$\uparrow$ & mATE$\downarrow$ &mASE$\downarrow$ & mAOE $\downarrow$ & mAVE $\downarrow$ &  mAAE $\downarrow$   & NDS $\uparrow$\\ 
\midrule

Baseline & 0.503 & 0.445 & 0.245 & 0.378 & 0.320 & 0.126 & 0.600  \\
Baseline + stereo input & 0.515 & 0.440 & 0.240 & 0.368 & 0.312 & 0.121 & 0.609\\
baseline + L1 &  0.514 & 0.442  & 0.238 & 0.365 & 0.313  & 0.123 & 0.609\\
Baseline + stereo input \& L1 & \textbf{0.521} & \textbf{0.437} & \textbf{0.235} & \textbf{0.359}  & \textbf{0.307} & \textbf{0.118} & \textbf{0.619} \\

\bottomrule
\end{tabular}

\caption{\label{tab:detection}{\textbf{Comparison of 3D detection results between baseline and adding depth prior on the nuScenes test set~\cite{caesar2020nuscenes}.} Taking Bevdepth~\cite{li2023bevdepth} as the baseline, we add depth prior as input and supervision in the depth estimation branch, and verify the effectiveness of depth estimation through the results of 3D object detection.
}}

\end{table*}

\subsubsection{Efficiency of Our Method}

Table~\ref{tab:FPS} illustrates the efficiency metrics of our approach, indicating an increase in memory usage and a slight reduction in frames per second (FPS) when compared to the baseline. This trade-off is deemed acceptable in light of the significant reduction in error rates our method achieves. Notably, in comparison with other state-of-the-art (SOTA) techniques, our method stands out by striking a superior balance between error minimization, memory efficiency, and FPS enhancement.
For a fair and consistent comparison, all results presented in Table~\ref{tab:FPS} and throughout this paper are obtained under identical experimental settings. As a strategy to further enhance the efficiency of our approach without compromising its accuracy, we implement a preprocessing step where images are resized to half their original resolution to the stereo matching process. 

\begin{table}[t!]
\centering
\renewcommand\arraystretch{1.5}
\setlength\tabcolsep{15pt}
\begin{tabular}{l|c|c}
\hline
Depth Estimation & IoU (\%) & mIoU (\%) \\ 
\hline
Baseline   & 71.2 & 40.7 \\
 \quad + stereo input  &71.4 & 41.0 \\
 \quad + L1   &71.5  & 41.2\\
 \quad + stereo input \& L1 & \textbf{71.7} & \textbf{42.4}\\
\hline
\end{tabular}
\caption{\label{tab:occ results}{ Results of
3D Occupancy prediction on the Occ3D-nuScenes~\cite{tian2023occ3d} dataset.} We use Voxformer~\cite{li2023voxformer} as baseline and apply our proposed method in depth estimation branch network. The effectiveness of depth prediction is verified by observing the effect of occupancy prediction.}
\end{table}

\subsection{Impact of Depth Estimation on Downstream Tasks}

Depth estimation plays a crucial role in 3D perception, significantly influencing the performance of downstream tasks within the perception system. Numerous works in vision-based detection~\cite{li2023bevdepth,wang2019pseudo}, semantic occupancy prediction~\cite{li2023voxformer,pan2023renderocc}, and bird's-eye-view representations~\cite{philion2020lift} integrate a depth estimation network as a foundational component. These studies collectively affirm that the precision of depth estimation directly correlates with the performance of subsequent tasks. Thus, evaluating the performance of downstream tasks under various depth estimation frameworks provides a tangible measure of depth estimation effectiveness.

\textbf{Depth Estimation's Influence on 3D Object Detection}

BevDepth~\cite{li2023bevdepth} illustrates the profound impact of employing an explicit depth estimation strategy on 3D object detection accuracy. In BevDepth's architecture, a dedicated branch network undertakes depth estimation by back-projecting 2D object features into 3D space. We experiment with replacing this branch with alternative depth estimation approaches to assess their performance impact.

Specificly, following the operation in BevDepth~\cite{li2023bevdepth}, we employ a pretrained VoVNet~\cite{lee2019energy} as the backbone on the nuScenes dataset~\cite{caesar2020nuscenes}, and construct a 3D object detection baseline. The depth estimation branch calculates the cross-entropy loss between the predicted depth and ground-truth depth  at a resolution of 640 $\times$ 1600. By applying our proposed depth prior, $\mathbf{D}_{p}$, derived from RAFT-Stereo~\cite{lipson2021raft} with optimized camera poses, we observe notable improvements. As Table~\ref{tab:detection} demonstrates, integrating the depth prior as input and supervision boosts the NDS from 0.600 to 0.619, surpassing the baseline. This gain is attributed not only to the prior's accuracy but also to its role in mitigating perspective inconsistencies.

\textbf{Depth Estimation's Impact on 3D Semantic Occupancy Prediction}

Further, we examine how varying depth estimation approaches affect 3D semantic occupancy prediction. Choosing VoxFormer~\cite{li2023voxformer} as our baseline, which employs a conventional depth estimation method for 2D-to-3D feature back-projection, we assess the impact of our proposed depth estimation technique on occupancy prediction performance. Implementing our method within VoxFormer without altering its network structure or parameters led to significant advancements.

Table~\ref{tab:occ results} reveals that incorporating the depth prior into both the input and supervision stages of depth estimation enhances occupancy prediction effectiveness, with mIoU scores improving from 40.7 to as high as 42.4. Figure~\ref{fig:occ} showcases that, compared to the baseline, the inclusion of depth priors yields more precise voxel occupancy and superior detection of smaller objects, primarily due to the enhanced geometric accuracy provided by the depth prior.

\section{Conclusion}
In this paper, we introduce a universal Stereo Guided Depth Estimation (SGDE) pipeline for 
\360 fisheye and pinhole camera sets. 
We propose to build virtual pinhole cameras  to unify the process of the two types of camera models.
We suggest forming the geometry loop constraint to calibrate camera poses, which significantly improves the accuracy and robustness of camera calibration in \360 views with minor overlap.
We demonstrate that explicitly rectifying and stereo matching adjacent images give us accurate depth prior. 
The depth prior can serve as the input and supervisor to the depth estimation model, which can consistently enhance the accuracy of all existing schemes and the cross-view geometric consistency.
Finally, we quantitatively and qualitatively demonstrate the effectiveness of our method on public datasets Synthetic Urban~\cite{won2019sweepnet}, DDAD~\cite{guizilini20203d} and nuScenes~\cite{caesar2020nuscenes}.  As a fundamental task in a 3D perception system, we highlight that the proposed depth estimation scheme can significantly improve downstream tasks and prove its contribution to the perception system.

\bibliographystyle{IEEEtran}
\bibliography{reference}

\end{document}